\documentclass{article}

\usepackage{arxiv}

\usepackage[utf8]{inputenc}
\usepackage[T1]{fontenc}
\usepackage[scaled=.98]{XCharter}
\usepackage[scaled=.95]{helvet}
\usepackage[scaled=1.1]{zlmtt}
\usepackage{amsmath}
\usepackage{amssymb}
\usepackage[uprightscript,charter,vvarbb,scaled=1.05]{newtxmath}
\usepackage[hyphens]{url}
\usepackage[table]{xcolor}
\definecolor{LinkBlue}{HTML}{2457C5}
\definecolor{AbstractGray}{HTML}{F3F5F7}
\definecolor{AbstractBorder}{HTML}{DDE3EA}
\usepackage[
  colorlinks=true,
  linkcolor=LinkBlue,
  citecolor=LinkBlue,
  urlcolor=LinkBlue
]{hyperref}
\usepackage{graphicx}
\usepackage{fontawesome5}
\usepackage{booktabs}
\usepackage{tabularx}
\usepackage{microtype}
\usepackage{enumitem}
\usepackage{etoolbox}
\usepackage[numbers,sort&compress]{natbib}
\usepackage{doi}
\usepackage[most]{tcolorbox}

\newtcolorbox{preprintabstract}{
  enhanced,
  breakable,
  colback=AbstractGray,
  colframe=AbstractBorder,
  boxrule=0.35pt,
  arc=3pt,
  left=12pt,
  right=12pt,
  top=5pt,
  bottom=9pt,
  before skip=10pt,
  after skip=16pt,
  title={Abstract},
  fonttitle=\large\bfseries\sffamily,
  coltitle=black,
  colbacktitle=AbstractGray,
  halign title=center,
  titlerule=0pt
}

\renewenvironment{abstract}
  {\begin{preprintabstract}\fontsize{10pt}{12.4pt}\selectfont}
  {\end{preprintabstract}}

\makeatletter
\renewenvironment{table}
  {\@float{table}}
  {\end@float}
\makeatother

\pdftrailerid{}

\newcommand{\NProtocolRescNoneSZeroFinal}{10.61}
\newcommand{\NProtocolRescNoneSOneFinal}{9.86}
\newcommand{\NProtocolRescLrdropSZeroFinal}{6.65}
\newcommand{\NProtocolRollbackEightZeroZeroSSevenFinal}{8.86}
\newcommand{\NProtocolRollbackOneZeroFiveZeroSSevenFinal}{9.40}
\newcommand{\NRopeRopeResumeMonSZeroFinal}{3.45}
\newcommand{\NRopeRopeResumeNoneSZeroFinal}{11.21}
\newcommand{\NRopeReplayCrossThreeZero}{6150}
\newcommand{\NRopeLeadSteps}{1500}
\newcommand{\NRopeLossAtCrossing}{3.47}
\newcommand{\NScaleThreeFiveZeroGptTwomRescNoneSZeroFinal}{8.28}
\newcommand{\NScaleThreeFiveZeroGptTwomRescLogitSZeroFinal}{4.04}
\newcommand{\NScaleThreeFiveZeroGptTwomRescFpThreeTwoSZeroFinal}{4.03}
\newcommand{\NScaleThreeFiveZeroGptTwomRescFpThreeTwoSZeroFinalVal}{4.02}
\newcommand{\NScaleThreeFiveZeroGptTwomRescFpThreeTwoSZeroFireStep}{700}
\newcommand{\NScaleThreeFiveZeroGptTwomRescFpThreeTwoSZeroFireValue}{278}
\newcommand{\NTriggerSgdDirtyLrOneemOneSZeroMaxSigma}{10.9}
\newcommand{\NAccReimplCrashedCount}{6}
\newcommand{\NAccReimplCrashEarliest}{700}
\newcommand{\NAccReimplCrashLatest}{1400}
\newcommand{\NOptSgdDirtyLrOneemOneSZeroCrash}{1700}
\newcommand{\NOptSgdDirtyLrThreeemTwoSZeroCrash}{3200}
\newcommand{\NOptSgdCleanLrOneemOneSZeroStableTo}{3900}
\newcommand{\NOptSgdCleanLrOneemOneSZeroFinal}{4.59}
\newcommand{\NOptSgdCleanLrThreeemTwoSZeroStableTo}{3900}
\newcommand{\NOptSgdCleanLrThreeemTwoSZeroFinal}{5.65}
\newcommand{\NFixQknormLongSZeroStableTo}{19900}
\newcommand{\NArchLlamaDirtyLongSZeroCrash}{13700}
\newcommand{\NArchLlamaDirtyLongSOneCrash}{13300}
\newcommand{\NArchGptTwoRopeDirtySZeroCrash}{7700}
\newcommand{\NArchLlamaNoropeDirtySZeroCrash}{1900}
\newcommand{\NArchLlamaNoropeDirtySOneCrash}{1400}
\newcommand{\NArchGptTwoRmsnormDirtySZeroCrash}{1400}
\newcommand{\NContinuumInjLamFourZeroZeroSZeroCrash}{2800}
\newcommand{\NContinuumInjLamTwoZeroZeroSZeroCrash}{5300}
\newcommand{\NContinuumInjLongBwdSZeroCrash}{6800}
\newcommand{\NContinuumInjLamZeroFiveZeroXTwentykSZeroCrash}{10100}
\newcommand{\NLocusStatsrouteNoneSZeroCrash}{2600}
\newcommand{\NLocusStatsrouteMlpSZeroCrash}{3000}
\newcommand{\NLocusStatsrouteQkSZeroStableTo}{7200}
\newcommand{\NLocusCorrectRopeMlpXOneThreekSZeroFinal}{4.50}
\newcommand{\NLocusCorrectGptTwomQkSZeroFinal}{3.88}
\newcommand{\NLocusCorrectLlamaQkSZeroFinal}{3.23}
\newcommand{\NLocusCorrectLlamaAllattnSZeroFinal}{3.23}
\newcommand{\NAnchorBiasonlyBwdSZeroStableTo}{9900}
\newcommand{\NAnchorBiasonlyBwdSZeroFinal}{3.31}
\newcommand{\NAnchorBiasonlyBwdSOneStableTo}{8600}
\newcommand{\NAnchorBiasonlyBwdSOneFinal}{3.40}
\newcommand{\NAnchorMeancorrectBwdSZeroCrash}{9200}
\newcommand{\NAnchorMeancorrectBwdSZeroMaxSigma}{76.8}
\newcommand{\NAnchorMeancorrectBwdSOneCrash}{5100}
\newcommand{\NAnchorMeancorrectHeadSZeroStableTo}{9900}
\newcommand{\NAnchorBiasonlyHeadSZeroCrash}{6000}
\newcommand{\NAnchorBiasonlyHeadSOneCrash}{5300}
\newcommand{\NAnchorBiasonlyHeadSTwoStableTo}{9900}
\newcommand{\NAnchorBiasonlyHeadSThreeStableTo}{9800}
\newcommand{\NFixSignflipLongSZeroStableTo}{8900}
\newcommand{\NFixSignflipLongSZeroFinal}{3.21}
\newcommand{\NFixSignflipLongSOneFinal}{3.25}
\newcommand{\NProtocolRescQkclipSZeroFinal}{3.67}
\newcommand{\NProtocolRescQkclipSOneFinal}{3.70}
\newcommand{\NProtocolRescDynmaxSZeroFinal}{6.45}
\newcommand{\NProtocolRescDynmaxSOneFinal}{8.07}
\newcommand{\NAccStatsBfOneSixSZeroBCrash}{2300}
\newcommand{\NAccStatsBfOneSixSOneBCrash}{2600}
\newcommand{\NAccObfOneSixStatsThreeTwoSZeroCrash}{1300}
\newcommand{\NAccObfOneSixStatsThreeTwoSOneCrash}{700}
\newcommand{\NForkActualCv}{0.210}
\newcommand{\NForkActualCvNull}{0.068}
\newcommand{\NForkActualMaxSigma}{52.0}
\newcommand{\NForkScramCv}{0.027}
\newcommand{\NForkScramStableTo}{3500}
\newcommand{\NForkHeadonlyCv}{0.204}
\newcommand{\NForkHeadonlyCvNull}{0.063}
\newcommand{\NForkHeadonlyMaxSigma}{62.7}
\newcommand{\NForkHeadcorrCv}{0.033}
\newcommand{\NEtwoMxfpFourSZeroMaxSigma}{292}
\newcommand{\NEtwoMxfpFourSZeroMaxLogit}{1.73e+06}
\newcommand{\NEtwoMxfpFourSOneMaxSigma}{205}
\newcommand{\NEtwoMxfpFourQknSZeroStableTo}{7900}
\newcommand{\NEtwoMxfpFourQknSZeroFinalVal}{3.39}
\newcommand{\NEtwoMxfpFourQknSZeroMaxSigma}{24}
\newcommand{\NEtwoMxfpFourQknSOneFinalVal}{3.41}
\newcommand{\NEtwoMxfpFourQknSOneMaxSigma}{26}
\newcommand{\NRopeRopeFullCtrlSZeroBFireStep}{8700}
\newcommand{\NRopeRopeFullCtrlSZeroBStableTo}{9450}
\newcommand{\NRopeRopeFullCtrlSZeroBFinalVal}{3.45}
\newcommand{\NRopeRopeFullCtrlSZeroBMaxLogit}{31.8}
\newcommand{\NRopeRopeFullNorescSZeroCrash}{6800}
\newcommand{\NRopeRopeFullNorescSZeroFinalVal}{11.29}
\newcommand{\NAccFlashFpThreeTwoaccSZeroBStableTo}{7900}
\newcommand{\NEdiagReimplLTwoRelDwq}{0.734}
\newcommand{\NEdiagSdpaLTwoRelDwq}{0.0164}
\newcommand{\NEdiagFoilLTwoRelDwq}{0.734}
\newcommand{\NEdiagReimplLinfDwq}{0.814}
\newcommand{\NEdiagFoilLinfDwq}{0.814}
\newcommand{\NFixBhSignflipSZeroStableTo}{8900}
\newcommand{\NEthreeMxfpFourSTwoCrash}{8500}
\newcommand{\NEthreeRescSZeroQknHundredFireStep}{4900}
\newcommand{\NEthreeRescSTwoQknHundredFireStep}{5300}
\newcommand{\NEthreeRescSTwoQknHundredFireLogit}{114}
\newcommand{\NEthreeRescSTwoQknHundredFinalVal}{3.19}
\newcommand{\NEthreeRescSTwoClipHundredFireStep}{4900}
\newcommand{\NEthreeRescSTwoQknThousandFireStep}{5900}
\newcommand{\NEthreeRescSTwoQknThousandFireLogit}{1656}
\newcommand{\NEthreeRescSTwoQknThousandStableTo}{19900}
\newcommand{\NEthreeRescSTwoQknThousandFinalVal}{3.19}
\newcommand{\NEthreeRescSTwoQknThousandPrefireSigma}{23}
\newcommand{\NEthreeRescSOneQknHundredFireStep}{5400}
\newcommand{\NEthreeRescSTwoQknThirtyFireStep}{4700}
\newcommand{\NEthreeRescSTwoQknThirtyFireLogit}{32}
\newcommand{\NEthreeRescSTwoQknThirtyStableTo}{18300}
\newcommand{\NEthreeRescSTwoQknThirtyFinalVal}{3.21}
\newcommand{\NEthreeRescSZeroClipHundredFireStep}{4600}
\newcommand{\NEthreeQuantAllCrashedCount}{11}
\newcommand{\NEthreeQknAllCrashedCount}{0}
\newcommand{\NIntervNofixCrashedCount}{4}
\newcommand{\NIntervQknCrashedCount}{0}
\newcommand{\NIntervMlpQknSZeroStable}{19900}
\newcommand{\NIntervVQknSZeroMaxSigma}{30}
\newcommand{\NIntervMlpQknSOneFinalVal}{3.24}
\newcommand{\NCapModeidSZeroOverlap}{0.40}
\newcommand{\NCapModeidSOneOverlap}{0.32}
\newcommand{\NDyadRefAllExtMaxSigma}{141}
\newcommand{\NDyadRefQkExtMaxSigma}{72}
\newcommand{\NEToENofixMaxSigma}{39}
\newcommand{\NEToENofixStableTo}{11900}
\newcommand{\NDyadNtwoAllTopkSZeroStable}{15900}
\newcommand{\NDyadNtwoAllTopkSOneMaxSigma}{15.1}
\newcommand{\NDyadNtwoQkTopkSZeroMaxSigma}{11.5}
\newcommand{\NDyadNtwoQkTopkSOneStable}{15200}
\newcommand{\NDyadNtwoAllRandkSZeroMaxSigma}{425}
\newcommand{\NDyadNtwoAllRandkSZeroCrash}{13200}
\newcommand{\NDyadNtwoAllRandkSOneCrash}{8300}
\newcommand{\NDyadNtwoQkRandkSZeroMaxSigma}{613}
\newcommand{\NBtwohundredQkNofixCrash}{5600}
\newcommand{\NBtwohundredMlpNofixCrash}{6000}
\newcommand{\NBtwohundredMlpRescueStableTo}{7800}
\newcommand{\NBtwohundredMlpRescueFinalVal}{3.22}
\newcommand{\NBtwohundredMlpRescueMaxSigma}{16}
\newcommand{\NBtwohundredMlpNofixEagerCrash}{5200}
\newcommand{\NBtwohundredMlpQkLogsigmaCorr}{0.98}
\newcommand{\NDyadNTwoAllShrinkSZeroCrash}{11600}
\newcommand{\NDyadNTwoQkShrinkSZeroCrash}{9000}
\newcommand{\NExpSufffullSThreeCrash}{11800}
\newcommand{\NExpSufffullSFiveCrash}{2500}
\newcommand{\NExpSufffullCrashedCount}{6}
\newcommand{\NExpNecSZeroCrash}{1100}
\newcommand{\NExpNecSOneCrash}{1900}
\newcommand{\NDyadNTwoAllToprenormSOneMaxSigma}{15.4}
\newcommand{\NDyadNTwoQkToprenormSZeroMaxSigma}{11.1}
\newcommand{\NDyadNTwoQkToprenormSZeroStable}{15900}
\newcommand{\NDyadNTwoQkToprenormSZeroFinalVal}{3.26}
\newcommand{\NDyadNTwoQkToprenormSOneFinalVal}{3.25}
\newcommand{\NCtrlAllCtrlSZeroFireStep}{4600}
\newcommand{\NCtrlGapSZeroFinalVal}{+0.0064}
\newcommand{\NCtrlAllCtrlSOneFireStep}{5600}
\newcommand{\NCtrlGapSOneFinalVal}{+0.0119}
\newcommand{\NDyadNTwoAllOfftgtrenormTwoSZeroCrash}{8100}
\newcommand{\NDyadNTwoAllOfftgtrenormTwoSOneCrash}{15600}
\newcommand{\NDyadNTwoAllOfftgtrenormTwoSOneMaxSigma}{491}
\newcommand{\NDyadNTwoQkOfftgtrenormTwoSZeroNoCrashTo}{15900}
\newcommand{\NDyadNTwoQkOfftgtrenormTwoSZeroMaxSigma}{237}
\newcommand{\NHorizonLhMargin}{8.32}
\newcommand{\NHorizonLhSurvivorPeakSigma}{46.5}
\newcommand{\NHorizonLhDeployfixPeakSigma}{46.5}
\newcommand{\NHorizonLhQknStableTo}{59900}
\newcommand{\NHorizonLhQknFinal}{3.13}
\newcommand{\NHorizonLhCtrlStableTo}{59900}
\newcommand{\NHorizonLhCtrlFinal}{3.13}
\newcommand{\NHorizonLhQknSOneFinal}{3.09}
\newcommand{\NHorizonLhQknSTwoFinal}{3.09}
\newcommand{\NHorizonLhCtrlSOneFinal}{3.10}
\newcommand{\NHorizonLhCtrlSTwoFinal}{3.09}
\newcommand{\NHorizonLhNofixSZeroCrash}{7200}
\newcommand{\NHorizonLhNofixSOneCrash}{8600}
\newcommand{\NHorizonLhNofixSTwoCrash}{7400}
\newcommand{\NDyadThreeFiftyQkToprenormSZeroMaxSigma}{19.0}
\newcommand{\NDyadThreeFiftyQkToprenormSZeroStable}{19900}
\newcommand{\NDyadThreeFiftyQkToprenormSOneMaxSigma}{22.0}
\newcommand{\NDyadThreeFiftyQkOfftgtrenormSZeroCrash}{16700}
\newcommand{\NDyadThreeFiftyQkOfftgtrenormSOneCrash}{13000}
\newcommand{\NDyadNTwoQkOfftgtThreeTwoKSTwoCrash}{13000}
\newcommand{\NDyadNTwoQkOfftgtThreeTwoKSThreeCrash}{27000}
\newcommand{\NDyadNTwoQkOfftgtThreeTwoKCrashedCount}{4}
\newcommand{\NArchLlamaMsDirtyCrashed}{4}
\newcommand{\NArchLlamaMsQknCrashed}{0}
\newcommand{\NLhdurMxfpFourQknStableTo}{199900}
\newcommand{\NLhdurMxfpFourQknMaxSigma}{65.1}
\newcommand{\NLhdurMxfpFourCtrlStableTo}{199900}
\newcommand{\NLhdurMxfpFourCtrlMaxSigma}{71.9}
\newcommand{\NLhaccFpThreeTwoaccStableTo}{103400}
\newcommand{\NLhaccNofixCrash}{700}
\newcommand{\NAccTOneFpThreeTwoaccSZeroStableTo}{4900}
\newcommand{\NAccTOneSdpaSZeroStableTo}{4900}
\newcommand{\NAccTOneSdpaflashSZeroStableTo}{4900}
\newcommand{\NAccTOneStandardSZeroStableTo}{4900}
\newcommand{\NAccTOneGptTwomSZeroStableTo}{4900}
\newcommand{\NAccTOneGptTwomRepoSZeroCrash}{900}
\newcommand{\NAccTOneGptTwomRepoSOneCrash}{1400}
\newcommand{\NStructureSFourActualLOneSZeroCrash}{1000}
\newcommand{\NStructureSFourActualLZeroFiveSZeroCrash}{1800}
\newcommand{\NStructureSFourActualLZeroFiveSOneCrash}{2000}
\newcommand{\NStructureSFourActualLZeroFiveSTwoCrash}{1600}
\newcommand{\NStructureSFourScramLOneSZeroStableTo}{7900}
\newcommand{\NStructureMatchedMagCoherentCrashed}{5}
\newcommand{\NStructureMatchedMagScrambleCrashed}{0}
\newcommand{\NStructureSFourBatchpermLOneSThreeCrash}{1000}
\newcommand{\NStructureSFourActualLOneSFiveCrash}{2900}
\newcommand{\NStructureInjBothRmsnoiseSTwoStableTo}{11900}
\newcommand{\NStructureBatchpermCrashed}{3}
\newcommand{\NStructureSFourBatchpermLOneSFourCrash}{1300}
\newcommand{\NStructureSFourScramLOneSTwoStableTo}{19900}
\newcommand{\NStructureSFourScramLOneSTwoFinal}{3.33}
\newcommand{\NStructureSFourScramLOneSThreeFinal}{3.36}
\newcommand{\NLocusStatsrouteNoneSOnecCrash}{2400}
\newcommand{\NLocusStatsrouteMlpSOnecCrash}{1900}
\newcommand{\NLocusStatsrouteQkSOnecStableTo}{11900}
\newcommand{\NOptSgdDirtyLrThreeemTwoSOneCrash}{2500}
\newcommand{\NScaleThreeFiveZeroGThreeFiveZeroRescNoneSZeroCrash}{900}
\newcommand{\NScaleThreeFiveZeroGThreeFiveZeroRescQknSZeroStableTo}{2900}
\newcommand{\NScaleThreeFiveZeroGThreeFiveZeroRescQknSZeroFireStep}{600}
\newcommand{\NAccTOneSdpaSOneOneZerokStableTo}{9900}
\newcommand{\NAccTOneSdpaSOneOneZerokFinal}{3.58}
\newcommand{\NScaleThreeFiveZeroGThreeFiveZeroRescNoneSThreeCrash}{1800}
\newcommand{\NScaleThreeFiveZeroGThreeFiveZerobRescNoneSZeroCrash}{2000}
\newcommand{\NScaleThreeFiveZeroGThreeFiveZerobRescQknSZeroStableTo}{2900}
\newcommand{\NScaleThreeFiveZeroGThreeFiveZerobRescQknSZeroFireStep}{700}
\newcommand{\NScaleThreeFiveZeroGThreeFiveZerobRescNoneSOneCrash}{1400}
\newcommand{\NScaleThreeFiveZeroGThreeFiveZeroNoneCrashed}{4}
\newcommand{\NScaleThreeFiveZeroGThreeFiveZeroRescuedCrashed}{0}
\newcommand{\NScaleThreeFiveZeroGThreeFiveZerobNoneCrashed}{2}
\newcommand{\NScaleThreeFiveZeroGThreeFiveZerobRescuedCrashed}{0}
\newcommand{\NRopeRopeThreeCtrlSOneStableTo}{15900}
\newcommand{\NRopeRopeThreeCtrlSOneFireStep}{13400}
\newcommand{\NRopeRopeThreeNorescSOneCrash}{13900}
\newcommand{\NRopeRopeThreeNorescSThreeCrash}{7400}
\newcommand{\NRopeRopeThreeCtrlSFourFireStep}{6100}
\newcommand{\NRopeRopeThreeCtrlFired}{6}
\newcommand{\NRopeRopeThreeCtrlCrashed}{0}
\newcommand{\NRopeRopeThreeNorescCrashed}{6}
\newcommand{\NAccTOneSdpaSOneOneZerokFinalVal}{3.60}
\newcommand{\NAccTOneSdpaCtrlSOneOneZerokFireStep}{5000}
\newcommand{\NAccTOneSdpaCtrlSOneOneZerokFinal}{3.35}
\newcommand{\NAccTOneSdpaCtrlSOneOneZerokFinalVal}{3.37}
\newcommand{\NAccTOneSdpaQknSOnebOneZerokFinal}{3.34}
\newcommand{\NAccTOneSdpaQknSOnebOneZerokFinalVal}{3.35}
\newcommand{\NLocusCorrectGptTwoQkSZeroStableTo}{3400}
\newcommand{\NLocusCorrectGptTwoVoSZeroStableTo}{1200}
\newcommand{\NLocusCorrectGptTwomQkSZeroStableTo}{2800}
\newcommand{\NLocusCorrectLlamaQkSZeroStableTo}{24800}
\newcommand{\NLocusCorrectRopeMlpXOneThreekSZeroStableTo}{12800}
\newcommand{\NLocusCorrectRopeQkXOneThreekSZeroStableTo}{12800}
\newcommand{\NRrOneCalTTenSZeroFireStep}{600}
\newcommand{\NRrOneCalTTenSZeroFireValue}{17.9}
\newcommand{\NRrOneCalTTenSZeroStableTo}{3850}
\newcommand{\NRrOneCalTTenSZeroFinalVal}{3.56}
\newcommand{\NRrOneCalTTenSOneFireStep}{600}
\newcommand{\NRrOneCalTTenSOneFireValue}{89.5}
\newcommand{\NRrOneCalTTenSOneStableTo}{3900}
\newcommand{\NRrOneCalTTenSOneFinalVal}{3.65}
\newcommand{\NRrOneCalTTwentySZeroFireStep}{650}
\newcommand{\NRrOneCalTTwentySZeroFireValue}{306.0}
\newcommand{\NRrOneCalTTwentySZeroStableTo}{3900}
\newcommand{\NRrOneCalTTwentySZeroFinalVal}{3.60}
\newcommand{\NRrOneCalTTwentySOneFireStep}{600}
\newcommand{\NRrOneCalTTwentySOneFireValue}{89.5}
\newcommand{\NRrOneCalTTwentySOneStableTo}{3900}
\newcommand{\NRrOneCalTTwentySOneFinalVal}{3.65}
\newcommand{\NRrOneCalTThirtySZeroFireStep}{650}
\newcommand{\NRrOneCalTThirtySZeroFireValue}{306.0}
\newcommand{\NRrOneCalTThirtySZeroStableTo}{3900}
\newcommand{\NRrOneCalTThirtySZeroFinalVal}{3.60}
\newcommand{\NRrOneCalTThirtySOneFireStep}{600}
\newcommand{\NRrOneCalTThirtySOneFireValue}{89.5}
\newcommand{\NRrOneCalTThirtySOneStableTo}{3850}
\newcommand{\NRrOneCalTThirtySOneFinalVal}{3.59}
\newcommand{\NRrOneCalTFiftySZeroFireStep}{650}
\newcommand{\NRrOneCalTFiftySZeroFireValue}{306.0}
\newcommand{\NRrOneCalTFiftySZeroStableTo}{3900}
\newcommand{\NRrOneCalTFiftySZeroFinalVal}{3.60}
\newcommand{\NRrOneCalTFiftySOneFireStep}{600}
\newcommand{\NRrOneCalTFiftySOneFireValue}{89.5}
\newcommand{\NRrOneCalTFiftySOneStableTo}{3900}
\newcommand{\NRrOneCalTFiftySOneFinalVal}{3.65}
\newcommand{\NRrOneCalTHundredSZeroFireStep}{650}
\newcommand{\NRrOneCalTHundredSZeroFireValue}{306.0}
\newcommand{\NRrOneCalTHundredSZeroStableTo}{3900}
\newcommand{\NRrOneCalTHundredSZeroFinalVal}{3.60}
\newcommand{\NRrOneCalTHundredSOneFireStep}{650}
\newcommand{\NRrOneCalTHundredSOneFireValue}{620.0}
\newcommand{\NRrOneCalTHundredSOneStableTo}{3900}
\newcommand{\NRrOneCalTHundredSOneFinalVal}{3.66}
\newcommand{\NRrOneCalNofixSZeroCrash}{1200}
\newcommand{\NRrOneCalNofixSZeroFinalVal}{9.25}
\newcommand{\NRrOneCalNofixSOneCrash}{700}
\newcommand{\NRrOneCalNofixSOneFinalVal}{10.82}
\newcommand{\NRrOneExtTTenSZeroFireStep}{1600}
\newcommand{\NRrOneExtTTenSZeroFireValue}{10.4}
\newcommand{\NRrOneExtTTenSZeroStableTo}{15900}
\newcommand{\NRrOneExtTTenSZeroFinalVal}{3.26}
\newcommand{\NRrOneExtTTenSOneFireStep}{1400}
\newcommand{\NRrOneExtTTenSOneFireValue}{10.7}
\newcommand{\NRrOneExtTTenSOneStableTo}{15900}
\newcommand{\NRrOneExtTTenSOneFinalVal}{3.26}
\newcommand{\NRrOneExtTTwentySZeroFireStep}{4800}
\newcommand{\NRrOneExtTTwentySZeroFireValue}{21.4}
\newcommand{\NRrOneExtTTwentySZeroStableTo}{15900}
\newcommand{\NRrOneExtTTwentySZeroFinalVal}{3.26}
\newcommand{\NRrOneExtTTwentySOneFireStep}{5200}
\newcommand{\NRrOneExtTTwentySOneFireValue}{21.2}
\newcommand{\NRrOneExtTTwentySOneStableTo}{15900}
\newcommand{\NRrOneExtTTwentySOneFinalVal}{3.27}
\newcommand{\NRrOneExtTFiftySZeroFireStep}{4700}
\newcommand{\NRrOneExtTFiftySZeroFireValue}{85.0}
\newcommand{\NRrOneExtTFiftySZeroStableTo}{15900}
\newcommand{\NRrOneExtTFiftySZeroFinalVal}{3.26}
\newcommand{\NRrOneExtTFiftySOneFireStep}{5300}
\newcommand{\NRrOneExtTFiftySOneFireValue}{66.0}
\newcommand{\NRrOneExtTFiftySOneStableTo}{15900}
\newcommand{\NRrOneExtTFiftySOneFinalVal}{3.27}
\newcommand{\NRrOneExtTHundredSZeroFireStep}{4800}
\newcommand{\NRrOneExtTHundredSZeroFireValue}{228.0}
\newcommand{\NRrOneExtTHundredSZeroStableTo}{15900}
\newcommand{\NRrOneExtTHundredSZeroFinalVal}{3.26}
\newcommand{\NRrOneExtTHundredSOneFireStep}{5500}
\newcommand{\NRrOneExtTHundredSOneFireValue}{102.0}
\newcommand{\NRrOneExtTHundredSOneStableTo}{15900}
\newcommand{\NRrOneExtTHundredSOneFinalVal}{3.27}
\newcommand{\NActScapThirtySZeroFireStep}{650}
\newcommand{\NActScapThirtySZeroFireValue}{306.0}
\newcommand{\NActScapThirtySZeroStableTo}{4950}
\newcommand{\NActScapThirtySOneFireStep}{600}
\newcommand{\NActScapThirtySOneFireValue}{89.5}
\newcommand{\NActScapThirtySOneStableTo}{4950}
\newcommand{\NActScapThirtySTwoFireStep}{650}
\newcommand{\NActScapThirtySTwoFireValue}{440.0}
\newcommand{\NActScapThirtySTwoStableTo}{4950}
\newcommand{\NActScapThirtySThreeFireStep}{600}
\newcommand{\NActScapThirtySThreeFireValue}{290.0}
\newcommand{\NActScapThirtySThreeStableTo}{4950}
\newcommand{\NActQkclipSZeroFireStep}{650}
\newcommand{\NActQkclipSZeroFireValue}{306.0}
\newcommand{\NActQkclipSZeroStableTo}{4950}
\newcommand{\NActQkclipSOneFireStep}{600}
\newcommand{\NActQkclipSOneFireValue}{89.5}
\newcommand{\NActQkclipSOneStableTo}{4950}
\newcommand{\NActQkclipSTwoFireStep}{650}
\newcommand{\NActQkclipSTwoFireValue}{440.0}
\newcommand{\NActQkclipSTwoStableTo}{4950}
\newcommand{\NActQkclipSThreeFireStep}{600}
\newcommand{\NActQkclipSThreeFireValue}{290.0}
\newcommand{\NActQkclipSThreeStableTo}{4950}
\newcommand{\NActScapEightSZeroFireStep}{650}
\newcommand{\NActScapEightSZeroFireValue}{306.0}
\newcommand{\NActScapEightSZeroStableTo}{4950}
\newcommand{\NActScapEightSOneFireStep}{600}
\newcommand{\NActScapEightSOneFireValue}{89.5}
\newcommand{\NActScapEightSOneStableTo}{4950}
\newcommand{\NActScapEightSTwoFireStep}{650}
\newcommand{\NActScapEightSTwoFireValue}{440.0}
\newcommand{\NActScapEightSTwoStableTo}{4950}
\newcommand{\NActScapEightSThreeFireStep}{600}
\newcommand{\NActScapEightSThreeFireValue}{290.0}
\newcommand{\NActScapEightSThreeStableTo}{4950}
\newcommand{\NActDynmaxSZeroFireStep}{650}
\newcommand{\NActDynmaxSZeroFireValue}{306.0}
\newcommand{\NActDynmaxSZeroCrash}{2000}
\newcommand{\NActDynmaxSOneFireStep}{600}
\newcommand{\NActDynmaxSOneFireValue}{89.5}
\newcommand{\NActDynmaxSOneCrash}{1400}
\newcommand{\NActDynmaxSTwoFireStep}{650}
\newcommand{\NActDynmaxSTwoFireValue}{440.0}
\newcommand{\NActDynmaxSTwoCrash}{2550}
\newcommand{\NActDynmaxSThreeFireStep}{600}
\newcommand{\NActDynmaxSThreeFireValue}{290.0}
\newcommand{\NActDynmaxSThreeCrash}{1950}
\newcommand{\NActScapsteThirtySZeroFireStep}{650}
\newcommand{\NActScapsteThirtySZeroFireValue}{306.0}
\newcommand{\NActScapsteThirtySZeroCrash}{1250}
\newcommand{\NActScapsteThirtySOneFireStep}{600}
\newcommand{\NActScapsteThirtySOneFireValue}{89.5}
\newcommand{\NActScapsteThirtySOneCrash}{850}
\newcommand{\NActScapsteThirtySTwoFireStep}{650}
\newcommand{\NActScapsteThirtySTwoFireValue}{440.0}
\newcommand{\NActScapsteThirtySTwoCrash}{1000}
\newcommand{\NActScapsteThirtySThreeFireStep}{600}
\newcommand{\NActScapsteThirtySThreeFireValue}{290.0}
\newcommand{\NActScapsteThirtySThreeCrash}{800}
\newcommand{\NActNofixSTwoCrash}{850}
\newcommand{\NActNofixSThreeCrash}{1500}
\newcommand{\NActQknThirtySTwoFireStep}{650}
\newcommand{\NActQknThirtySTwoFireValue}{440.0}
\newcommand{\NActQknThirtySTwoStableTo}{4950}
\newcommand{\NActQknThirtySThreeFireStep}{600}
\newcommand{\NActQknThirtySThreeFireValue}{290.0}
\newcommand{\NActQknThirtySThreeStableTo}{4950}
\newcommand{\NActHPrimarySZero}{3900}
\newcommand{\NActScapThirtySZeroValH}{3.58}
\newcommand{\NActScapThirtySZeroValRefH}{3.60}
\newcommand{\NActScapThirtySZeroGapH}{-0.0177}
\newcommand{\NActQkclipSZeroValH}{3.60}
\newcommand{\NActQkclipSZeroGapH}{+0.0018}
\newcommand{\NActHPrimarySOne}{3850}
\newcommand{\NActScapThirtySOneValH}{3.58}
\newcommand{\NActScapThirtySOneValRefH}{3.59}
\newcommand{\NActScapThirtySOneGapH}{-0.0091}
\newcommand{\NActQkclipSOneValH}{3.59}
\newcommand{\NActQkclipSOneGapH}{+0.0013}
\newcommand{\NActHPrimarySTwo}{4950}
\newcommand{\NActScapThirtySTwoValH}{3.50}
\newcommand{\NActScapThirtySTwoValRefH}{3.52}
\newcommand{\NActScapThirtySTwoGapH}{-0.0182}
\newcommand{\NActQkclipSTwoValH}{3.51}
\newcommand{\NActQkclipSTwoGapH}{-0.0064}
\newcommand{\NActHPrimarySThree}{4950}
\newcommand{\NActScapThirtySThreeValH}{3.51}
\newcommand{\NActScapThirtySThreeValRefH}{3.52}
\newcommand{\NActScapThirtySThreeGapH}{-0.0163}
\newcommand{\NActQkclipSThreeValH}{3.52}
\newcommand{\NActQkclipSThreeGapH}{-0.0000}
\newcommand{\NActScapEightSZeroValH}{3.59}
\newcommand{\NActScapEightSZeroGapH}{-0.0074}
\newcommand{\NActScapEightSOneValH}{3.59}
\newcommand{\NActScapEightSOneGapH}{-0.0038}
\newcommand{\NActScapEightSTwoValH}{3.50}
\newcommand{\NActScapEightSTwoGapH}{-0.0160}
\newcommand{\NActScapEightSThreeValH}{3.51}
\newcommand{\NActScapEightSThreeGapH}{-0.0144}
\newcommand{\NActScapsteThirtySZeroValH}{11.22}
\newcommand{\NActScapsteThirtySZeroGapH}{+7.6249}
\newcommand{\NActScapsteThirtySOneValH}{8.93}
\newcommand{\NActScapsteThirtySOneGapH}{+5.3425}
\newcommand{\NActScapsteThirtySTwoValH}{11.32}
\newcommand{\NActScapsteThirtySTwoGapH}{+7.8089}
\newcommand{\NActScapsteThirtySThreeValH}{11.05}
\newcommand{\NActScapsteThirtySThreeGapH}{+7.5284}
\newcommand{\NActDynmaxSZeroValH}{8.54}
\newcommand{\NActDynmaxSZeroGapH}{+4.9467}
\newcommand{\NActDynmaxSOneValH}{8.35}
\newcommand{\NActDynmaxSOneGapH}{+4.7645}
\newcommand{\NActDynmaxSTwoValH}{9.37}
\newcommand{\NActDynmaxSTwoGapH}{+5.8545}
\newcommand{\NActDynmaxSThreeValH}{8.16}
\newcommand{\NActDynmaxSThreeGapH}{+4.6425}
\newcommand{\NRrTwoQkSTwoCrash}{7900}
\newcommand{\NRrTwoQkqknSTwoStableTo}{19900}
\newcommand{\NRrTwoMlpSTwoCrash}{7000}
\newcommand{\NRrTwoMlpqknSTwoStableTo}{19900}
\newcommand{\NRrTwoVSTwoCrash}{7400}
\newcommand{\NRrTwoVqknSTwoStableTo}{19900}

\usepackage{array}
\usepackage{caption}
\usepackage{placeins}
\makeatletter
\renewenvironment{table*}{\@float{table}}{\end@float}
\renewenvironment{figure*}{\@float{figure}}{\end@float}
\makeatother

\definecolor{tabink}{HTML}{1E2B38}      %
\definecolor{tabaccent}{HTML}{2B4C6F}   %
\definecolor{tabheadbg}{HTML}{E1E8F0}   %
\definecolor{tabgroupbg}{HTML}{EEF2F7}  %
\definecolor{tabzebra}{HTML}{F7F9FB}    %
\newcolumntype{Y}{>{\raggedright\arraybackslash}X}
\newcolumntype{P}[1]{>{\raggedright\arraybackslash}p{#1}}
\newcolumntype{Q}[1]{>{\centering\arraybackslash}p{#1}}
\newcolumntype{R}[1]{>{\raggedleft\arraybackslash}p{#1}}
\newcommand{\tabstubw}{0.46\textwidth}
\newcommand{\tabcollw}{0.075\textwidth}
\newlength{\tabpad}
\newlength{\tabgapskip}
\newcommand{\tabsetup}{\normalsize\setlength{\tabcolsep}{5pt}\renewcommand{\arraystretch}{1.0}%
  \setlength{\tabgapskip}{4pt}\setlength{\heavyrulewidth}{0.9pt}\setlength{\lightrulewidth}{0.4pt}%
  \setlength{\cmidrulewidth}{0.3pt}\setlength{\aboverulesep}{0pt}\setlength{\belowrulesep}{0pt}%
  \arrayrulecolor{tabink}\rowcolors{1}{}{tabzebra}}
\newcommand{\tabsetupdense}{\tabsetup\renewcommand{\arraystretch}{1.0}%
  \setlength{\tabgapskip}{2.5pt}}
\newcommand{\colhead}[1]{\leavevmode\textcolor{tabink}{\scshape #1}}%
\newcommand{\tabheadrow}{\rowcolor{tabheadbg}\noalign{\global\rownum=0 }}
\newcommand{\tabgroup}[2]{\rowcolor{tabgroupbg}\noalign{\global\rownum=0 }%
  \multicolumn{#1}{@{\hspace{\tabpad}}l}{\color{tabaccent}\itshape #2}}
\newcommand{\tabpanel}[2]{\rowcolor{white}\noalign{\global\rownum=0 }%
  \multicolumn{#1}{@{\hspace{\tabpad}}l}{\color{tabaccent}\bfseries #2}}
\newcommand{\tabgap}{\addlinespace[\tabgapskip]}
\newcommand{\tabpanelsep}{\midrule\addlinespace[4pt]}
\makeatletter
\newcommand{\tabsign}[1]{\edef\tab@n{#1}\expandafter\tab@sign\tab@n\@nil}
\def\tab@sign#1#2\@nil{\ifx-#1\textminus#2\else#1#2\fi}
\makeatother
\newcommand{\arxivlayout}{}
\newcommand{\papercolumnfigurewidth}{\textwidth}
\newcommand{\maintextname}{main text}
\newcommand{\paperpartscope}{main text and appendix alike}
\newcommand{\artifactname}{public v2 artifact}

\newcommand{\paperstatus}{Preprint}
\newcommand{\dezhicorrespondenceemail}{dezhiran@pku.edu.cn}
\newcommand{\taocorrespondenceemail}{taoxie@pku.edu.cn}
\title{\normalfont\sffamily\bfseries One QK Channel, Many Sources:\\Tracing Low-Precision Attention Collapse}

\newcommand{\artifacturl}{https://github.com/xieTwim/one-qk-channel-artifact/tree/v0.2.0-arxiv-v2}
\newcommand{\artifactdisplay}{github.com/xieTwim/one-qk-channel-artifact}
\newcommand{\authornamefont}{\fontsize{11.8pt}{14.2pt}\selectfont\sffamily\bfseries}
\newcommand{\affiliationfont}{\fontsize{10.2pt}{12.4pt}\selectfont\normalfont}
\newlength{\authorrowbreakheight}
\makeatletter
\patchcmd{\@maketitle}
  {\begin{tabular}[t]{c}\bf\rule{\z@}{24\p@}\ignorespaces}
  {\begin{tabular}[t]{c}\authornamefont\rule{\z@}{24\p@}\ignorespaces}
  {}{\PackageWarning{arxiv-paper-template}{Could not patch the first author-row break}}
\patchcmd{\@maketitle}
  {\begin{tabular}[t]{c}\bf\rule{\z@}{24\p@}\ignorespaces}
  {\begin{tabular}[t]{c}\authornamefont\rule{\z@}{\authorrowbreakheight}\ignorespaces}
  {}{\PackageWarning{arxiv-paper-template}{Could not patch the second author-row break}}
\makeatother
\newcommand{\resourcelink}[2]{%
  \href{#1}{\normalfont\mdseries\nolinkurl{#2}}%
}
\newcommand{\resourceentry}[4]{%
  \raisebox{-0.05ex}{\makebox[1.35em][c]{\normalsize #1}}&
  \textbf{#2:}\ \resourcelink{#3}{#4}%
}

\author{
  \authornamefont
  \textbf{Shuxiao Xie}\textsuperscript{1,2}\thanks{Equal contribution.}
  \quad
  \textbf{Shuyang Xie}\textsuperscript{3}\footnotemark[1]
  \quad
  \textbf{Yuan Cao}\textsuperscript{1,4}
  \AND
  \textbf{Dezhi Ran}\textsuperscript{1}\thanks{Corresponding authors:
  \resourcelink{mailto:\dezhicorrespondenceemail}{\dezhicorrespondenceemail} and
  \resourcelink{mailto:\taocorrespondenceemail}{\taocorrespondenceemail}.}
  \quad
  \textbf{Wei Yang}\textsuperscript{2}
  \quad
  \textbf{Tao Xie}\textsuperscript{1,2,4,5}\footnotemark[2]
  \\[0.7em]
  \affiliationfont
  \textsuperscript{1}Beijing Tongming Lake Information Technology Application
  Innovation Center (TLAIC), China
  \\
  \textsuperscript{2}Fudan University Institute of Systems for Advanced Computing, China
  \\
  \textsuperscript{3}Harbin Institute of Technology, China
  \\
  \textsuperscript{4}Key Lab of HCST (PKU), MOE; SCS, Peking University, Beijing, China
  \\
  \textsuperscript{5}Shanghai Institute of Systems for Open Computing, China
}

\date{\small
\begin{tabular}{@{}r@{\hspace{0.4em}}l@{}}
\resourceentry{\faGithub}{Research artifact (v2)}{\artifacturl}{\artifactdisplay}
\end{tabular}
}
\renewcommand{\headeright}{\paperstatus}
\renewcommand{\undertitle}{\paperstatus}
\renewcommand{\shorttitle}{One QK Channel, Many Sources}
\hypersetup{pdftitle={One QK Channel, Many Sources: Tracing Low-Precision Attention Collapse},pdfsubject={Public preprint},pdfauthor={Shuxiao Xie, Shuyang Xie, Yuan Cao, Dezhi Ran, Wei Yang, Tao Xie}}

\begin{document}
\maketitle

\begin{abstract}
A bfloat16 transformer can train normally for many steps, then collapse abruptly. Prior work links
the collapse to attention errors whose shared structure lets rounding contributions compound, and
shows QK normalization disrupts that compounding. Distinct low-precision errors trigger the same
collapse, leaving unclear whether each needs a fix at its source or one shared route can be blocked
instead. We isolated the fault behind a reproduced GPT-2-class collapse to the streaming-softmax
accumulator, where an fp32 streaming core repairs it, and turned it into an assay for moving a
controlled error across sources. Using it, we found that errors placed
outside attention still drove the same QK spectral runaway, and that correcting only QK kept
training stable while the fault stayed active. This is a \emph{source-channel dissociation}: fault
source is not failure channel. It held across tested architectures and scales, and reproduced on a
second GPU architecture.
As a causal probe, projecting each update off the current QK weights' leading three singular
directions held the query projection's largest singular value to
\NDyadNTwoQkToprenormSZeroMaxSigma{}, whereas removing equal energy elsewhere left it at
\NDyadNTwoQkOfftgtrenormTwoSZeroMaxSigma{}: the QK channel causally drives the early
runaway. What lets the injected error in is \emph{temporal sign-coherence}, its per-head sign persisting
across steps, not aggregate deviation; once inside, the runaway shows as attention-logit
saturation. If one channel carries these faults, a QK-local action switched on mid-run should
contain each. \emph{QK-Guard}, a dormant controller, tests this by switching on parameter-free QK
normalization at the first monitored threshold crossing. On the runs designated for this test, the QK-local action prevented the failure of each matched or
same-configuration unguarded run; on plain GPT-2, all 12 final train and validation losses were within
0.03 nat of same-configuration always-on QK-norm, and both methods ran 60k steps without collapse. Intervention at the QK locus therefore suffices in
place of a fix at each source.
\end{abstract}

\providecommand{\papercolumnfigurewidth}{\columnwidth}

\section{Introduction}

Bfloat16 transformer training is routine \citep{deepseekai-deepseekv3-2024}, yet a
run can appear healthy through long optimization before abruptly collapsing, revealing little about
what accumulated. Prior work links this pattern to structurally aligned attention errors with
compounding rounding contributions and identifies query--key (QK) normalization as what disrupts
that compounding \citep{qiu-why-2025}. The same normalization also prevents attention-logit growth, a known
instability of large-scale training \citep{dehghani-scaling-2023,wortsman-smallscale-2023}.

The open question is whether low-precision faults require repair at their sources or instead share a
blockable failure channel; collapse alone cannot decide, since it reveals only the endpoint. We therefore pair controlled fault-source moves with always-on module-routed gradient correction in an
identification assay that varies the fault source independently of the route receiving a clean fp32 gradient,
with the rest fixed and each source mechanism left enabled, so the splice tests channel blocking while every source stays unrepaired. Faults within and outside attention converge on
the same QK spectral runaway, and correcting only the QK route blocks that channel. This
source-channel dissociation is our thesis: fault source is not failure
channel~\ifhmode\unskip\fi.

Together, these results change what counts as a sufficient remedy: because heterogeneous sources share the
QK channel, one action at the QK locus can replace a separate repair at each source. An fp32
streaming core fixes the streaming-softmax accumulator fault but not post-reduction
weight-gradient quantization or errors outside attention. We test that implication in flight with
QK-Guard, which watches the downstream attention-logit saturation because ordinary feature learning is
temporally coherent too.

Our contributions are three findings about the QK channel, each turning on a discriminating contrast.

\begin{enumerate}
    \item \textbf{A source-channel dissociation.} We pair controlled fault-source moves with always-on
    module-routed gradient correction while leaving each source mechanism enabled: faults originating within
    and outside attention converge on the same dominant QK channel, which correction on the QK route blocks and
    correction on the MLP route does not.

    \item \textbf{The entry condition for the QK channel.} We manipulate error-sign sequences to identify
    temporal sign-coherence (the persistence of each head's error sign across optimizer steps) as what
    decides whether an injected error enters the QK channel, separating it from the per-tile
    scalar mean and aggregate magnitude that prior analysis leaves coupled to it; against an equal-energy
    off-target control, removing each run's own dominant QK weight directions verifies that the channel causally drives the early runaway.

    \item \textbf{Containing the channel in flight.} QK-Guard, a dormant closed loop, tests whether acting at
    the QK locus mid-run contains these faults: it watches attention-logit saturation and, when its monitored
    statistic first crosses the threshold, applies parameter-free per-head QK normalization. Across the
    runs designated for this test it prevents the failure of each matched or same-configuration unguarded run without
    repairing the fault or rewinding the run, and at the same trigger a softmax-local dynamic maximum and a
    learning-rate cut fail, so reactivity alone does not determine rescue.
\end{enumerate}

\section{Isolating a Fault and Building the Assay}

To edit one numerical factor at a time, we reproduced the GPT-2-class collapse reported by
\citet{qiu-why-2025} on GPT-2 small \citep{radford-gpt2-2019} trained on OpenWebText
\citep{gokaslan-openwebtext-2019}, in a PyTorch re-implementation of FlashAttention-2's tiled
online-softmax recurrence \citep{dao-flashattention2-2023} derived from that report's public code. Its
carries are the running output \(O\) and softmax statistics (stats), the running maximum and exponential
sum \citep{milakov-online-2018}. Vendor kernels keep these carries in fp32; this
implementation stores them in bf16 and exposes the recurrence for controlled edits, which lets a fault stay
active while one route at a time is cleaned. Table~\ref{tab:localization} groups the edits by the uncertainty
removed.
The kernel's maintainer had publicly flagged accumulator precision as one untested implementation
difference \citep{dao-flashattention-discussion-1931}. The streaming core's precision is central: three protocols observe no collapse once the
same tiled implementation runs its core in fp32: that arm at two seeds, the fp32/fp32 carry condition at two seeds, and one long run
manually stopped while healthy at \NLhaccFpThreeTwoaccStableTo{} logged steps.  The fp32-core
intervention keeps tiling on while changing the outcome, so tiling is not the factor; the full-matrix
arm corroborates. Together with panel (b), where either bf16 carry alone suffices to collapse, these contrasts
localize collapse to a bf16 streaming-accumulator artifact
rather than bf16 operands, the FlashAttention kernel family, tiling, or a single-scale
artifact~\ifhmode\unskip\fi.

\begin{table}[!tb]
\centering
\caption{Localizing the collapse. Each row changes the factor it names; the streaming-core row keeps tiling on
while the core's precision changes. ``Coll.'' is collapsed over tested runs under the sustained-divergence rule
(Appendix~A); \(\times\) collapse, \(\blacktriangleright\) a last observed step without collapse (an observation
bound). SDPA is PyTorch's fused path. In panel (b), both collapsing carries show the same runaway in
\(\sigma(W^Q)\), the query projection's largest singular value.}
\label{tab:localization}
\ifdefined\arxivlayout
\tabsetup
\begin{tabularx}{\textwidth}{@{\hspace{\tabpad}}Y Q{0.13\textwidth} P{0.2\textwidth}@{\hspace{\tabpad}}}
\toprule
\tabpanel{3}{(a)~Localizing the fault}\\
\tabheadrow\colhead{Condition} & \colhead{Coll.} & \colhead{Step}\\
\midrule
\tabgroup{3}{The object, and the fp32-core intervention on it}\\
bf16 $O$ + bf16 stats & \NAccReimplCrashedCount{}/\NAccReimplCrashedCount{}
  & $\times$\,\NAccReimplCrashEarliest{}--\NAccReimplCrashLatest{} \\
\;\;streaming core $\to$ fp32 & 0/2
  & $\blacktriangleright$\,\NAccTOneFpThreeTwoaccSZeroStableTo{} \\
\tabgap
\tabgroup{3}{Scope: is another factor sufficient?}\\
SDPA, bf16 operands & 0/2 & $\blacktriangleright$\,\NAccTOneSdpaSZeroStableTo{} \\
FlashAttention-2 forced & 0/2 & $\blacktriangleright$\,\NAccTOneSdpaflashSZeroStableTo{} \\
Full matrix, no tiling & 0/2 & $\blacktriangleright$\,\NAccTOneStandardSZeroStableTo{} \\
\tabgap
\tabgroup{3}{Scale: GPT-2 medium}\\
re-implementation & 2/2
  & $\times$\,\NAccTOneGptTwomRepoSZeroCrash{}, \NAccTOneGptTwomRepoSOneCrash{} \\
SDPA & 0/2 & $\blacktriangleright$\,\NAccTOneGptTwomSZeroStableTo{} \\
\tabpanelsep
\tabpanel{3}{(b)~Which streaming carry: three observed conditions}\\
\tabheadrow\colhead{$O$ / stats dtype} & \colhead{Coll.} & \colhead{Step}\\
\midrule
bf16 / fp32 & 2/2
  & $\times$\,\NAccObfOneSixStatsThreeTwoSOneCrash{}, \NAccObfOneSixStatsThreeTwoSZeroCrash{} \\
fp32 / bf16 & 2/2
  & $\times$\,\NAccStatsBfOneSixSZeroBCrash{}, \NAccStatsBfOneSixSOneBCrash{} \\
fp32 / fp32 & 0/2 & $\blacktriangleright$\,\NAccFlashFpThreeTwoaccSZeroBStableTo{} \\
\bottomrule
\end{tabularx}
\else
\footnotesize
\setlength{\tabcolsep}{4pt}
\begin{tabular}{@{}l c l@{}}
\toprule
\multicolumn{3}{@{}l@{}}{\textbf{(a)~Localizing the fault}}\\
\cmidrule(r){1-3}
Condition & Coll. & Step\\
\midrule
\multicolumn{3}{@{}l@{}}{\itshape The object, and the fp32-core intervention on it}\\
bf16 $O$ + bf16 stats & \NAccReimplCrashedCount{}/\NAccReimplCrashedCount{}
  & $\times$\,\NAccReimplCrashEarliest{}--\NAccReimplCrashLatest{} \\
\;\;streaming core $\to$ fp32 & 0/2
  & $\blacktriangleright$\,\NAccTOneFpThreeTwoaccSZeroStableTo{} \\
\addlinespace[2pt]
\multicolumn{3}{@{}l@{}}{\itshape Scope: is another factor sufficient?}\\
SDPA, bf16 operands & 0/2 & $\blacktriangleright$\,\NAccTOneSdpaSZeroStableTo{} \\
FlashAttention-2 forced & 0/2 & $\blacktriangleright$\,\NAccTOneSdpaflashSZeroStableTo{} \\
Full matrix, no tiling & 0/2 & $\blacktriangleright$\,\NAccTOneStandardSZeroStableTo{} \\
\addlinespace[2pt]
\multicolumn{3}{@{}l@{}}{\itshape Scale: GPT-2 medium}\\
re-implementation & 2/2
  & $\times$\,\NAccTOneGptTwomRepoSZeroCrash{}, \NAccTOneGptTwomRepoSOneCrash{} \\
SDPA & 0/2 & $\blacktriangleright$\,\NAccTOneGptTwomSZeroStableTo{} \\
\addlinespace[3pt]
\multicolumn{3}{@{}l@{}}{\textbf{(b)~Which streaming carry: three observed conditions}}\\
\cmidrule(r){1-3}
$O$ / stats dtype & Coll. & Step\\
\midrule
bf16 / fp32 & 2/2
  & $\times$\,\NAccObfOneSixStatsThreeTwoSOneCrash{}, \NAccObfOneSixStatsThreeTwoSZeroCrash{} \\
fp32 / bf16 & 2/2
  & $\times$\,\NAccStatsBfOneSixSZeroBCrash{}, \NAccStatsBfOneSixSOneBCrash{} \\
fp32 / fp32 & 0/2 & $\blacktriangleright$\,\NAccFlashFpThreeTwoaccSZeroBStableTo{} \\
\bottomrule
\end{tabular}
\fi
\end{table}

The two carries enter backward differently. The output carry reaches \(dQ\) and \(dK\), never \(dV\): saved \(O\) re-enters only through the
\(D\)-term \(D_i=dO_i^{\top}O_i\), the per-row inner product of the output gradient with \(O\), which enters the score
gradient as \(dS_{ij}=P_{ij}(dP_{ij}-D_i)\)~\citep{dao-flashattention-2022} (four-bit attention training keeps a high-precision output for this
term alone~\citep{zhang-attnqat-2026}). The statistics carry reaches \(dQ\), \(dK\), and \(dV\) through the saved
log normalizer. Either carry produces the same \(\sigma(W^Q)\) runaway; the split decomposes one
streaming-accumulator source with its generator fixed and leaves open whether other sources converge on it.

\paragraph{A manipulable assay.}
Only the output carry yields an exactly computable injection: its bf16-minus-fp32 saved-output difference
enters through the per-row \(D\)-term while an fp32-core forward pass keeps the residual stream clean, and it
is recomputed each step from the current \(Q\), \(K\), and \(V\), so the error tracks current weights and data
rather than a stored tensor. That makes the error movable along four axes: source, placement (an emulation sub-choice within a source), dose, and structure. Scaling between the
fp32-core and bf16-accumulation endpoints turns the same-code contrast into a dose continuum.

Under backward-only placement, reduced doses delayed collapse monotonically to
\NContinuumInjLamFourZeroZeroSZeroCrash{}, \NContinuumInjLamTwoZeroZeroSZeroCrash{},
\NContinuumInjLongBwdSZeroCrash{}, and \NContinuumInjLamZeroFiveZeroXTwentykSZeroCrash{}, in that
order (Appendix~D). Over a pre-registered window of healthy-phase states and eight batches, the median one-step relative \(\ell_2\) error in \(dW^Q\) against an fp32 reference was \NEdiagReimplLTwoRelDwq{} under the bf16-accumulation
re-implementation and \NEdiagSdpaLTwoRelDwq{} under fp32-accumulating SDPA. In one seed, a matched-recipe fp32-core arm was stopped
while still healthy at \NLhaccFpThreeTwoaccStableTo{} steps, whereas its untreated control collapsed at
\NLhaccNofixCrash{}. An fp32 streaming core is therefore effectively
immune~\ifhmode\unskip\fi.

The assay supports two tests: \S3 replaces bf16 accumulation with post-reduction weight-gradient quantization
across module targets inside and outside attention, everything else fixed, after first locating where the
fault must be intercepted; \S4 applies matched-magnitude structure transforms with everything else fixed.

\section{Identifying a Common QK Failure Channel Across Fault Sources}
\label{sec:identifying}

To test whether repair must act where error originates, we invert two manipulations. Route correction
keeps \S2's accumulator fault active while moving the repair through a clean-gradient splice; source
shift fixes the repair while replacing bf16 accumulation with a different error at a different site; together
they test whether heterogeneous generators converge on one query--key channel that one action blocks while
each generator remains active.

\subsection{Locating the blocking channel}

Each step runs the same weights and batch backward through the faulty re-implementation and the
fp32-core reference, splicing the clean gradient into one named route while leaving all others
faulty. Each row therefore tests whether its named route must be intercepted;
Table~\ref{tab:locus} shows that the anticipated architecture-specific rescue does not occur.

Cleaning only the query--key route rescues every tested architecture---plain GPT-2, a RoPE variant
\citep{su-roformer-2021}, a 350M model, and a LLaMA-style architecture \citep{touvron-llama-2023}---to train
losses from \NLocusCorrectLlamaQkSZeroFinal{} to \NLocusCorrectGptTwomQkSZeroFinal{}; MLP cleaning rescues none, and on LLaMA whole-attention cleaning reaches the same train loss as the query--key route
(\NLocusCorrectLlamaAllattnSZeroFinal{} in both), placing the effective position inside that block. A four-seed
LLaMA check without splicing gives \NArchLlamaMsDirtyCrashed{} of four untreated collapses and
\NArchLlamaMsQknCrashed{} of four under per-head query--key normalization~\citep{henry-querykey-2020}. Moving
the fault to the kernel's other carry preserves the split: bf16 softmax statistics, the path that also reaches \(dV\), collapse two untreated route-grid seeds at \NLocusStatsrouteNoneSZeroCrash{} and
\NLocusStatsrouteNoneSOnecCrash{}; query--key cleaning keeps both healthy through
\NLocusStatsrouteQkSZeroStableTo{} and \NLocusStatsrouteQkSOnecStableTo{} logged steps, whereas MLP cleaning
collapses at \NLocusStatsrouteMlpSZeroCrash{} and \NLocusStatsrouteMlpSOnecCrash{}. Thus query--key correction is
the route-level blocking intervention across the tested grid~\ifhmode\unskip\fi.

\subsection{Moving the fault source}

Because every prior arm shares one generator, QK's blocking role could be specific to bf16 accumulation, so we
fix the intervention and change the generator: the second one rounds the materialized weight gradient after
backward and reduction, before clipping and the optimizer step, onto an MXFP4-style
grid~\citep{rouhani-microscaling-2023} of four-bit elements in blocks of thirty-two sharing a power-of-two scale. Attention itself runs unchanged under bf16 autocast through the fused SDPA path
(bf16 operands, fp32 accumulation) in every arm, so the selected module gradient is the only injected
low-precision source; attention's own rounding error is present in all arms, including the QK-normalized twins.

Applied to the query--key gradient alone, the quantizer drives the same spectral runaway in \(W^Q\), to
\NEtwoMxfpFourSZeroMaxSigma{} and \NEtwoMxfpFourSOneMaxSigma{}, with attention logits reaching
\NEtwoMxfpFourSZeroMaxLogit{}; per-head query--key normalization suppresses it to
\NEtwoMxfpFourQknSZeroMaxSigma{} and \NEtwoMxfpFourQknSOneMaxSigma{} at healthy validation losses of
\NEtwoMxfpFourQknSZeroFinalVal{} and \NEtwoMxfpFourQknSOneFinalVal{}. At a matched grid and magnitude,
deterministic and stochastic rounding~\citep{gupta-deep-2015} both produce the runaway signature~\ifhmode\unskip\fi; loss failure
within this shorter window is seed-variable, and the separately powered all-weight-gradient extension through 20k
records
\NEthreeQuantAllCrashedCount{} of 12 quantized collapses, versus
\NEthreeQknAllCrashedCount{} of 4 with query--key normalization.

Moving the identical emulator, block size, and per-step seed schedule to MLP and value gradients
yields \NIntervNofixCrashedCount{} of four unprotected collapses on two data-parallel ranks after the same \(W^Q\) runaway
(Figure~\ref{fig:dissociation}a; other rank counts in Appendix~D). Per-head query--key normalization leaves
\NIntervQknCrashedCount{} of four collapsing, holds the runaway near
\NIntervVQknSZeroMaxSigma{}, and brings both targets to a healthy
\NIntervMlpQknSOneFinalVal{} validation loss. Changing only the target, the matched sweep shows two generators reach one channel; one action
contains both. The weight-gradient quantizer spans four sites, three outside the query--key path.

Applying the same four-bit grid to the GEMM operands, with fp32 accumulation, leaves training healthy through
\NEToENofixStableTo{} steps with the spectral norm no higher than \NEToENofixMaxSigma{}: four-bit arithmetic is not
by itself the trigger in these configurations, an observed boundary at a milder dose (Appendix~D).

\begin{figure}[!t]
\centering
\includegraphics[width=\textwidth]{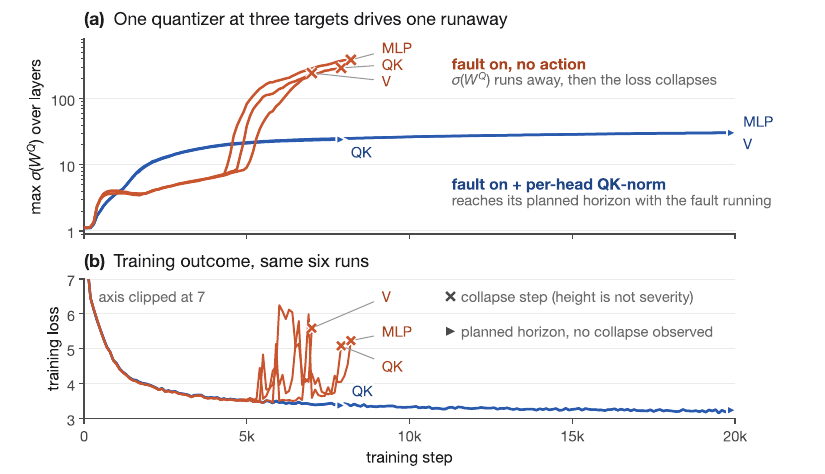}
\caption{Source-channel dissociation. \textbf{(a)}~one weight-gradient quantizer routed to the \(QK\), MLP
or value weight gradient drives the same \(\sigma(W^Q)\) runaway (one quantizer at three locations); per-head \(QK\)-norm twins survive with the fault running, the action capping the attention logit while leaving \(\sigma\) uncapped (Figure~\ref{fig:b200}). \textbf{(b)}~outcomes on the same step axis. Labels name the target at each run's end; one seed per arm, planned horizons
\NEtwoMxfpFourQknSZeroStableTo{} or \NIntervMlpQknSZeroStable{} steps; \(\times\) a collapse step (height is not
severity), \(\blacktriangleright\) a planned horizon.}
\label{fig:dissociation}
\end{figure}

\subsection{Portability and scope}

On a single Blackwell B200 running our exact training code, software-emulated quantizer included, the
dissociation reproduces: an
off-attention MLP source still develops the query--key runaway and collapses, while per-head query--key
normalization triggered mid-run holds a second run of the same faulted configuration healthy (Figure~\ref{fig:b200};
per-arm steps and the eager control in Appendix~D).

RoPE, RMSNorm and scale delay this failure without preventing it. The delay depends on RoPE, but no
combination of the three prevents the outcome; safety here comes from an fp32
attention core or from query--key
normalization~\ifhmode\unskip\fi. Table~\ref{tab:locus}(b) gives the ladder. The failure also appears
with SGD in place of AdamW: the untreated fault collapses while the matched clean controls do not
(Appendix~A).

The dissociation these results establish has a falsifiable bound: the query--key
channel is the dominant attractor for coherent low-precision error that perturbs residual-block
parameter updates or enters upstream of those blocks, because in this pre-norm setting the query--key
projections read a normalized input. The falsifier follows: a source acting after the final
attention module has no later query--key computation to be intercepted at, and would escape. Within
that scope every source we moved the fault to reached the same channel.
That scope leaves the gate unresolved: whether coherence or sheer magnitude opens the channel.

\ifdefined\arxivlayout
\begin{table}[!tbp]
\else
\begin{table}[!t]
\fi
\centering
\caption{Route correction across architectures. Panel (a): collapsed over seeds run; empty cells not run.
Panel (b): the same untreated fault per architecture, the last row adding per-head \(QK\)-norm to
RoPE\,+\,RMSNorm. \(^{\dagger}\)The non-collapsing seed plateaus at \NLocusCorrectRopeMlpXOneThreekSZeroFinal{} train loss through \NLocusCorrectRopeMlpXOneThreekSZeroStableTo{} steps, above every QK-corrected final (\NLocusCorrectLlamaQkSZeroFinal{}--\NLocusCorrectGptTwomQkSZeroFinal{}). \(^{\ddagger}\)Seed-variable (one RoPE seed healthy, one plain-GPT-2 run
truncated while diverging); no claim rests on this row.}
\label{tab:locus}
\ifdefined\arxivlayout
\tabsetup
\begin{tabularx}{\textwidth}{@{\hspace{\tabpad}}Y *{4}{Q{0.12\textwidth}}@{\hspace{\tabpad}}}
\toprule
\tabpanel{5}{(a)~Which route must be corrected?}\\
\tabheadrow\colhead{Gradient route cleaned} & \colhead{plain} & \colhead{+RoPE} & \colhead{350M} & \colhead{LLaMA}\\
\midrule
none (fault untreated)
  & 1/1 & 2/2 & 2/2 & 1/1 \\
\textbf{$QK$ only}
  & \textbf{0/2} & \textbf{0/3} & \textbf{0/2} & \textbf{0/1} \\
MLP only
  & 2/2 & 2/3\rlap{$^{\dagger}$} & 2/2 & 1/1 \\
$V$ and $O$
  & 1/1\rlap{$^{\ddagger}$} & 1/2\rlap{$^{\ddagger}$} & --- & 1/1 \\
whole attention block
  & --- & 0/1 & --- & 0/1 \\
\tabpanelsep
\tabpanel{5}{(b)~Architecture moves the timescale; \(QK\)-norm moves the outcome}\\
\tabheadrow\colhead{Configuration} & \multicolumn{4}{c@{\hspace{\tabpad}}}{\colhead{Collapsed, step}}\\
\midrule
RMSNorm alone & \multicolumn{4}{c@{\hspace{\tabpad}}}{\NArchGptTwoRmsnormDirtySZeroCrash{}} \\
RoPE alone & \multicolumn{4}{c@{\hspace{\tabpad}}}{\NArchGptTwoRopeDirtySZeroCrash{}} \\
RoPE + RMSNorm & \multicolumn{4}{c@{\hspace{\tabpad}}}{\NArchLlamaDirtyLongSOneCrash{} and \NArchLlamaDirtyLongSZeroCrash{}} \\
\;\;four further seeds & \multicolumn{4}{c@{\hspace{\tabpad}}}{\NArchLlamaMsDirtyCrashed{} of four} \\
\;\;minus RoPE & \multicolumn{4}{c@{\hspace{\tabpad}}}{\NArchLlamaNoropeDirtySOneCrash{} and \NArchLlamaNoropeDirtySZeroCrash{}} \\
\addlinespace[2pt]
\;\;+ $QK$-norm & \multicolumn{4}{c@{\hspace{\tabpad}}}{none of 4 seeds, to \NFixQknormLongSZeroStableTo{}} \\
\bottomrule
\end{tabularx}
\else
\footnotesize
\setlength{\tabcolsep}{3pt}
\begin{tabular}{@{}>{\raggedright\arraybackslash}p{0.30\columnwidth}
                  cccc@{}}
\toprule
\multicolumn{5}{@{}l@{}}{\textbf{(a)~Which route must be corrected?}}\\
\cmidrule(r){1-5}
Gradient route cleaned & plain & +RoPE & 350M & LLaMA\\
\midrule
none (fault untreated)
  & 1/1 & 2/2 & 2/2 & 1/1 \\
\textbf{$QK$ only}
  & \textbf{0/2} & \textbf{0/3} & \textbf{0/2} & \textbf{0/1} \\
MLP only
  & 2/2 & 2/3\rlap{$^{\dagger}$} & 2/2 & 1/1 \\
$V$ and $O$
  & 1/1\rlap{$^{\ddagger}$} & 1/2\rlap{$^{\ddagger}$} & --- & 1/1 \\
whole attention block
  & --- & 0/1 & --- & 0/1 \\
\addlinespace[3pt]
\multicolumn{5}{@{}l@{}}{\textbf{(b)~Architecture moves the timescale; \(QK\)-norm moves the outcome}}\\
\cmidrule(r){1-5}
Configuration & \multicolumn{4}{c}{Collapsed, step}\\
\midrule
RMSNorm alone
  & \multicolumn{4}{c}{\NArchGptTwoRmsnormDirtySZeroCrash{}} \\
RoPE alone
  & \multicolumn{4}{c}{\NArchGptTwoRopeDirtySZeroCrash{}} \\
RoPE + RMSNorm
  & \multicolumn{4}{c}{\NArchLlamaDirtyLongSOneCrash{} and \NArchLlamaDirtyLongSZeroCrash{}} \\
\;\;four further seeds
  & \multicolumn{4}{c}{\NArchLlamaMsDirtyCrashed{} of four} \\
\;\;minus RoPE
  & \multicolumn{4}{c}{\NArchLlamaNoropeDirtySOneCrash{} and \NArchLlamaNoropeDirtySZeroCrash{}} \\
\addlinespace[2pt]
\;\;+ $QK$-norm
  & \multicolumn{4}{c}{none of 4 seeds, to \NFixQknormLongSZeroStableTo{}} \\
\bottomrule
\end{tabular}
\fi
\end{table}

\section{Explaining Entry and Testing the Causal Role of the QK Channel}
\label{sec:explaining}

\S3 located where failure can be blocked; it left open what admits the large, structured errors observed there and whether the query--key channel's directions drive it. Structure interventions at matched magnitude, where possible, answer the first question; removing the channel's
dominant directions answers the second.
\subsection{Temporal sign-coherence, not aggregate deviation, gates entry}

We return to the injection assay of \S2 at one magnitude and one width, applying one predefined structure
transformation to the injected backward error per cell at matched per-injection magnitude. Left unaltered, the
error collapses training in
\NStructureMatchedMagCoherentCrashed{} of five seeds, between
\NStructureSFourActualLOneSZeroCrash{} and \NStructureSFourActualLOneSFiveCrash{} steps. Scrambling its sign element by element at the same magnitude collapses
\NStructureMatchedMagScrambleCrashed{} of four (two seeds censored at \NStructureSFourScramLOneSZeroStableTo{}
steps, two healthy at \NStructureSFourScramLOneSTwoStableTo{}, ending at \NStructureSFourScramLOneSTwoFinal{} and
\NStructureSFourScramLOneSThreeFinal{}). Permuting the error along the batch axis, which preserves the
batch-summed tensor at every head and position while reassigning values across examples, collapses
\NStructureBatchpermCrashed{} of three, between \NStructureSFourBatchpermLOneSThreeCrash{} and
\NStructureSFourBatchpermLOneSFourCrash{} steps. Gaussian noise at matched magnitude collapses
neither of two seeds through \NStructureInjBothRmsnoiseSTwoStableTo{} steps. Destroying which
example a contribution came from does not prevent the collapse; destroying the sign
does at the tested natural scale~\ifhmode\unskip\fi. Each alternative reading is
excluded by a different arm: batch permutation excludes example identity, the matched-magnitude scramble and
noise arms exclude magnitude, the mean-removal pair below excludes the scalar mean, and a smaller dose does not
rescue the coherent error, which at half the magnitude still collapses on three seeds in the natural context, at
\NStructureSFourActualLZeroFiveSTwoCrash{}, \NStructureSFourActualLZeroFiveSZeroCrash{} and
\NStructureSFourActualLZeroFiveSOneCrash{} steps.

A second pair separates two quantities that a prior derivation of this collapse leaves
coupled~\citep{qiu-why-2025}: a mean component of the backward error and the error's temporal structure.
Injecting only the error's scalar mean, reduced over each tile, is benign, holding two seeds to
\NAnchorBiasonlyBwdSZeroStableTo{} and \NAnchorBiasonlyBwdSOneStableTo{} steps at \NAnchorBiasonlyBwdSZeroFinal{}
and \NAnchorBiasonlyBwdSOneFinal{}, while the coherent error with that mean removed still collapses both seeds,
at \NAnchorMeancorrectBwdSOneCrash{} and \NAnchorMeancorrectBwdSZeroCrash{} steps through the same
\(\sigma(W^Q)\) runaway, reaching \NAnchorMeancorrectBwdSZeroMaxSigma{}. Neither aggregate magnitude nor that
scalar gates entry~\ifhmode\unskip\fi. The derivation
of~\citet{qiu-why-2025} reaches its conclusion through coherence and already names query--key normalization as
the disruptor; what this pair adds, for our tile-reduced scalar, is that the two quantities can be moved
independently (Appendix~C).

The mean matters one level down: per-head DC bias, the same reduction with the head axis retained, is a sufficient
carrier. Injected alone, reduced over the whole pass and across ranks in the natural low-precision forward, it
collapses all \NExpSufffullCrashedCount{} full-horizon seeds, from \NExpSufffullSFiveCrash{} to
\NExpSufffullSThreeCrash{}~\ifhmode\unskip\fi. Removing it protects two seeds through
\NAnchorMeancorrectHeadSZeroStableTo{} steps in the clean-forward surrogate (forward pass kept clean), while full-pass removal in the
natural forward still collapses at \NExpNecSZeroCrash{} and \NExpNecSOneCrash{}
steps~\ifhmode\unskip\fi: these assays show necessity in the clean-forward surrogate and
sufficiency in the natural forward, and a separate near-threshold per-tile per-head backward-only injection also collapses two of
four seeds.

Sign randomization protects under two complementary constructions: whole-tensor randomization holds both
arms to \NFixSignflipLongSZeroStableTo{} steps at \NFixSignflipLongSZeroFinal{} and
\NFixSignflipLongSOneFinal{}~\ifhmode\unskip\fi, and per-head signs drawn once per
optimizer step, which hold each head's sign fixed within the step and so keep its exact per-step carrier
magnitude, hold both seed-matched arms to the same horizon~\ifhmode\unskip\fi.
With the four matched sign-scramble survivors and three half-dose true-sign crashers, these rule out a
magnitude-only explanation and jointly identify temporal sign persistence as a load-bearing tested
discriminator in this regime (Table~\ref{tab:coherence}, Appendix~C).

\subsection{Direction-specific removal suppresses the QK runaway}

Everything so far changes what the channel receives; whether the channel itself drives the runaway is still
open. The last experiment removes the channel's own dominant directions: at each optimizer step we form the
three leading singular directions of the current \(W^Q\) and \(W^K\), project that step's realized update off
them, and rescale the remainder to the update's original norm, so the update loses its component along the
weight's leading directions at unchanged total energy. Against the weight-gradient sources of \S3.2 this
suppresses the runaway: all four treated arms hold \(\sigma(W^Q)\) between \NDyadNTwoQkToprenormSZeroMaxSigma{} and
\NDyadNTwoAllToprenormSOneMaxSigma{} and none collapses through
\NDyadNTwoQkToprenormSZeroStable{} steps, ending between
\NDyadNTwoQkToprenormSOneFinalVal{} and \NDyadNTwoQkToprenormSZeroFinalVal{} validation loss.

Suppression alone would not settle direction, because removing any energy from the update might do it. The
control removes equal energy from outside that subspace, leaving the target directions in place, at matched
per-step update norm and intervention magnitude, and does not suppress (Figure~\ref{fig:causal-removal}):
across the same four cells \(\sigma(W^Q)\) runs to between \NDyadNTwoQkOfftgtrenormTwoSZeroMaxSigma{} and
\NDyadNTwoAllOfftgtrenormTwoSOneMaxSigma{}, the two model-wide arms collapse at
\NDyadNTwoAllOfftgtrenormTwoSZeroCrash{} and \NDyadNTwoAllOfftgtrenormTwoSOneCrash{} steps, and the two
query--key arms carry the runaway without collapsing inside this schedule (0 of 2 through
\NDyadNTwoQkOfftgtrenormTwoSZeroNoCrashTo{} steps), with four fresh seeds of that cell on a longer schedule
collapsing \NDyadNTwoQkOfftgtThreeTwoKCrashedCount{} of four. Merely shrinking the update without targeting a
direction does not rescue either: all four such arms collapse, between \NDyadNTwoQkShrinkSZeroCrash{} and
\NDyadNTwoAllShrinkSZeroCrash{} steps, so net reduction of update energy is neither necessary nor sufficient.
The same specificity holds at 350M: on-target removal carries two seeds to
\NDyadThreeFiftyQkToprenormSZeroStable{} steps at \NDyadThreeFiftyQkToprenormSZeroMaxSigma{} and
\NDyadThreeFiftyQkToprenormSOneMaxSigma{}, while the equal-energy off-target pair collapses at
\NDyadThreeFiftyQkOfftgtrenormSOneCrash{} and \NDyadThreeFiftyQkOfftgtrenormSZeroCrash{} steps.

\begin{figure}[!t]
\centering
\includegraphics[width=\textwidth]{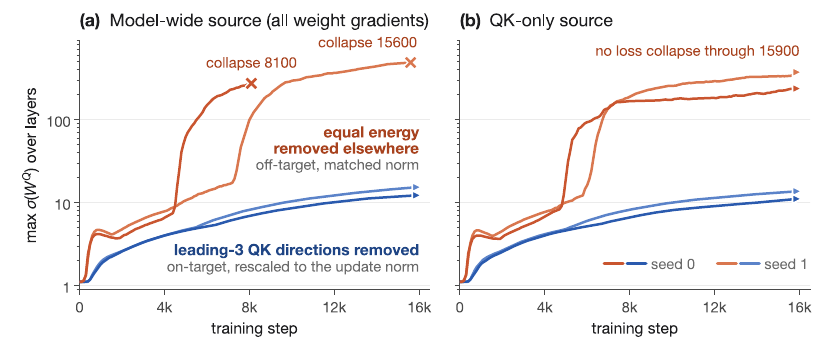}
\caption{Direction-specific removal suppresses the \(QK\) runaway. Each step's realized update is projected
off the three leading singular directions of the current \(W^Q\) and \(W^K\) and restored to its norm; the
control removes equal energy from outside that subspace at matched norm. On-target removal suppresses the
runaway under \textbf{(a)}~a model-wide and \textbf{(b)}~a \(QK\)-only source; the control does not. Two seeds per arm (seed 0 darker, seed 1 lighter shade); \(\times\) marks a collapse step; in (b) the
off-target arms run away without loss collapse inside this schedule (\(\blacktriangleright\) at
\NDyadNTwoQkOfftgtrenormTwoSZeroNoCrashTo{} steps).}
\label{fig:causal-removal}
\end{figure}

The query--key channel is therefore a causal driver of the early spectral runaway, and of the collapse that follows it inside these horizons. The removal serves as a causal probe; \S5 supplies the remedy. The sources share one query--key-local channel whose directions move with the training state: their runaway directions overlap only partially (Appendix~D).

\section{Closing the Channel in Flight}
\label{sec:closing}

\subsection{Monitor downstream, act at QK}

Acting at the QK locus in flight requires choosing when: cumulative QK update-coherence sits near
unity in doomed and safe runs alike, with overlapping ranges and even inverted minima, because ordinary
feature learning is temporally coherent too~\ifhmode\unskip\fi. The spectral norm the
runaway acts on is no better as a trigger, since healthy and doomed \(\sigma(W^Q)\) ranges also overlap
and an acceleration rule misses the spectrally flat RoPE collapse~\ifhmode\unskip\fi
(nor is high \(\sigma(W^Q)\) sufficient for loss failure at deployment length, Appendix~C). QK-Guard therefore
monitors the
runaway's downstream attention-logit signature, permanently switching on parameter-free per-head QK
normalization at the QK locus at its first monitored crossing of the fixed threshold of 30.
The statistic is, per layer, the mean over a fixed four-sequence validation probe of each query's largest
causal logit, maximized over layers, read every 50 to 200 steps (per arm family) on the post-update weights, so a fire is the first periodic reading above 30 and normalization is active from the
next step (Appendix~A). The value was fixed once, on the plain GPT-2 fault, before the reported guarded transfer runs were launched, and
never re-tuned; Appendix~A gives its provenance and a two-seed sweep from 10 to 100 on the plain fault and
the held-out family (every setting fires, and no guarded arm collapses through its last logged step while both
untreated plain seeds collapse). On the reference arms it was set on, it separates converged healthy envelopes
from doomed pre-collapse trajectories~\ifhmode\unskip\fi.
It serves as a trigger: a fire switches on the normalization that always-on training carries from step zero (Appendix~C reports a fire on fault-free
training). On plain GPT-2, both unguarded seeds end at \NProtocolRescNoneSZeroFinal{} and
\NProtocolRescNoneSOneFinal{} train loss, whereas all six guarded arms fire before collapse and finish
healthy~\ifhmode\unskip\fi (Figure~\ref{fig:guard}, which also shows \S5.2's same-fire-step
action contrast).

\begin{figure}[!t]
\centering
\includegraphics[width=\textwidth]{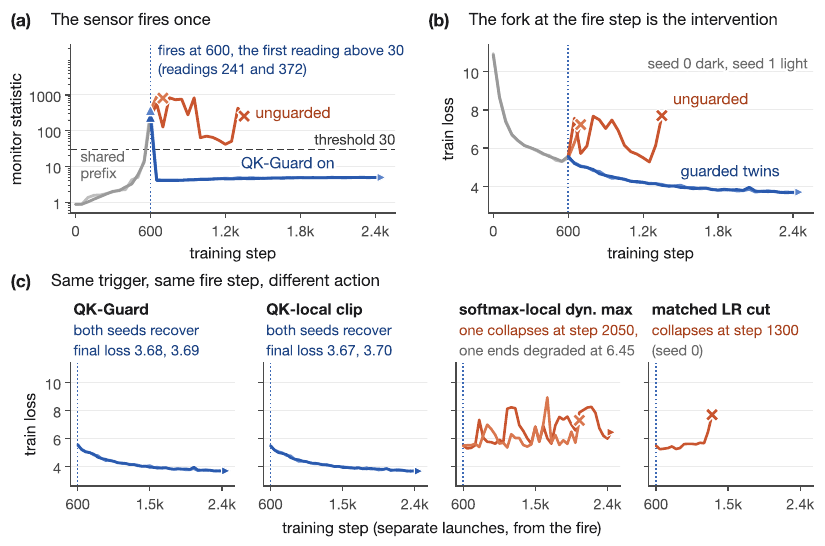}
\caption{QK-Guard anatomy. \textbf{(a)}~the monitor fires (\(\blacktriangle\)) at
the first periodic reading of the attention-logit statistic (Appendix~A) above the fixed threshold of 30
(dashed) and switches on per-head \(QK\) normalization for good; the reading already sits above the
threshold because monitoring is periodic. \textbf{(b)}~unguarded runs collapse (\(\times\)); their guarded
twins reach the planned horizon (\(\blacktriangleright\)) healthy; each seed-matched pair (seed 0 dark, seed 1 light) is
identical through the fire step (gray), the fork at the dotted line being the intervention. \textbf{(c)}~separately launched runs, shown from the fire step: at the same trigger and fire step, both \(QK\)-local
actions rescue both seeds, the softmax-local dynamic maximum fails (one collapse, one degraded endpoint;
\(\blacktriangleright\) marks censoring, never health), and the matched learning-rate cut still diverges
(\S5.2).}
\label{fig:guard}
\end{figure}

\subsection{At a fixed trigger, the action decides rescue}

A working alarm settles when to act; holding the trigger and the fire step fixed turns the choice of action into a controlled comparison. The QK-local clip, a second QK-local action distinct from the normalization that QK-Guard fires, restores healthy training on both seeds, at final train losses of \NProtocolRescQkclipSZeroFinal{} and
\NProtocolRescQkclipSOneFinal{}, while the softmax-local dynamic maximum does not, one seed plateauing degraded
at \NProtocolRescDynmaxSZeroFinal{} and the other collapsing at
\NProtocolRescDynmaxSOneFinal{}~\ifhmode\unskip\fi; a softmax-local cap at the clip's nominal bound
also rescues, but only with its derivative, which damps the gradient saturated scores send to \(Q\) and \(K\) (Appendix~C). The generic controls fail in different ways: a tenfold learning-rate cut at the same fire step still diverges (\NProtocolRescLrdropSZeroFinal{}),
rollback with reshuffling and an optimizer reset re-collapses from both post-onset restore points
(\NProtocolRollbackEightZeroZeroSSevenFinal{} and \NProtocolRollbackOneZeroFiveZeroSSevenFinal{}), and always-on
gradient clipping was already active in every collapsed run reported here~\ifhmode\unskip\fi.

Online, thresholded query--key intervention already exists: Kimi K2's query--key clip measures per-head
maximum logits, rescales query and key weights above a threshold, and reports an ablation of that threshold~\citep{team-kimi-2025},
and a recent characterization likewise intervenes mid-run~\citep{su-characterization-2025}. QK-Guard instead
permanently enables normalization after the first monitored crossing. Our evidence is comparative:
at the same trigger and fire step, the softmax-local dynamic maximum fails while the QK-local clip
rescues~\ifhmode\unskip\fi. The other interventions (Appendix~C) act from step zero or on
magnitude: always-on query--key normalization~\citep{henry-querykey-2020,dehghani-scaling-2023,olmo-2-2024},
loss scaling and rollback~\citep{micikevicius-mixed-2018,chowdhery-palm-2022}, gradient
clamping~\citep{zclip-2025,spam-2025,adagc-2025}, and spectral learning-rate control~\citep{qi-taming-2025};
detectors report without acting~\citep{huang-mechanismdriven-2026,qiu-spectral-2025}, and predictors
anticipate faults~\citep{wortsman-smallscale-2023,emadi-rankaware-2026}.

\subsection{Transfer Across Architecture, Scale, and Fault Source}

The threshold was fixed on the plain GPT-2 fault; RoPE permits a causal test of architecture
transfer. Switching on the monitor at a pre-collapse checkpoint on the doomed RoPE baseline's own trajectory
makes the guard fire while loss is still normal, and the run continues past the step where its monitor-off twin
collapses, ending at \NRopeRopeResumeMonSZeroFinal{} versus the
twin's \NRopeRopeResumeNoneSZeroFinal{}~\ifhmode\unskip\fi. Replaying that baseline
locates the crossing at step \NRopeReplayCrossThreeZero{}, at least \NRopeLeadSteps{} steps before
collapse, with loss still healthy at \NRopeLossAtCrossing{}~\ifhmode\unskip\fi.

Independent RoPE launches replicate beyond the replay: an independently seeded from-scratch control fires at
\NRopeRopeFullCtrlSZeroBFireStep{} (maximum logit \NRopeRopeFullCtrlSZeroBMaxLogit{}) and stays healthy to
\NRopeRopeFullCtrlSZeroBStableTo{} at \NRopeRopeFullCtrlSZeroBFinalVal{} validation loss, while the monitor-off
configuration, on its own trajectory, collapses at \NRopeRopeFullNorescSZeroCrash{} and ends at
\NRopeRopeFullNorescSZeroFinalVal{}~\ifhmode\unskip\fi;
a later pair, bitwise identical through the fire step, reproduces the split.
Six further matched 16000-step pairs (Table~\ref{tab:guard}) fire in \NRopeRopeThreeCtrlFired{} of six cases
between
\NRopeRopeThreeCtrlSFourFireStep{} and \NRopeRopeThreeCtrlSOneFireStep{};
\NRopeRopeThreeNorescCrashed{} of six unguarded twins collapse after their pair's fire step, whereas
\NRopeRopeThreeCtrlCrashed{} of six guarded arms collapse, and every guarded arm reaches
\NRopeRopeThreeCtrlSOneStableTo{} logged steps.

Scale transfer is tested at 350M with the same threshold: query--key normalization ends at
\NScaleThreeFiveZeroGptTwomRescLogitSZeroFinal{}, and an fp32 streaming-core hot-swap at
\NScaleThreeFiveZeroGptTwomRescFpThreeTwoSZeroFinal{} train and \NScaleThreeFiveZeroGptTwomRescFpThreeTwoSZeroFinalVal{}
validation loss, versus the shared unguarded arm's
\NScaleThreeFiveZeroGptTwomRescNoneSZeroFinal{}~\ifhmode\unskip\fi.
Six further seeds in a same-regime configuration replicate the pattern across both actions:
\NScaleThreeFiveZeroGThreeFiveZeroNoneCrashed{} of four untreated seeds collapse against
\NScaleThreeFiveZeroGThreeFiveZeroRescuedCrashed{} of eight rescued arms at twice the original batch,
\NScaleThreeFiveZeroGThreeFiveZerobNoneCrashed{} of two against \NScaleThreeFiveZeroGThreeFiveZerobRescuedCrashed{}
of four batch-matched (Appendix~C).

A held-out fault tests source transfer: quantizing every weight gradient after reduction replaces the tuning
fault entirely, yet threshold and action carry over unchanged: both seeds fire, at steps
\NCtrlAllCtrlSZeroFireStep{} and \NCtrlAllCtrlSOneFireStep{}, and finish within \NCtrlGapSZeroFinalVal{} and
\NCtrlGapSOneFinalVal{} nats of always-on query--key normalization, while one same-configuration reference run
collapses~\ifhmode\unskip\fi.

\subsection{Endpoint gaps and deployment durability}

\providecommand{\NHorizonLhDeployfixPeakSigma}{\NHorizonLhSurvivorPeakSigma}

Across the six plain rescue arms, every raw final train and validation endpoint gap to the always-on query--key normalization reference of the same configuration is below 0.03 nats~\ifhmode\unskip\fi. For
durability, three seeds per method run a 60k-step schedule: none of the six always-on or QK-Guard arms
collapses through \NHorizonLhQknStableTo{} logged steps (validation endpoints per seed in Appendix~C), whereas
the three controls collapse at \NHorizonLhNofixSZeroCrash{}, \NHorizonLhNofixSOneCrash{}, and
\NHorizonLhNofixSTwoCrash{}, so survivors run \NHorizonLhMargin{} times longer than the earliest collapse. Their peak \(\sigma(W^Q)\), \NHorizonLhDeployfixPeakSigma{}, far exceeds values at which unguarded runs
collapse, so durability is read from the loss. Under the model-wide four-bit weight-gradient fault, one seed per arm, always-on query--key normalization reaches \NLhdurMxfpFourQknStableTo{} logged steps of a 200k schedule with \(\sigma\) at \NLhdurMxfpFourQknMaxSigma{}, and QK-Guard reaches \NLhdurMxfpFourCtrlStableTo{} with \(\sigma\) at \NLhdurMxfpFourCtrlMaxSigma{}; neither collapses.

\section{Conclusion}

Low-precision faults with different sources (a bf16 streaming-softmax accumulator, and four-bit weight-gradient quantization applied inside or outside attention) converge on one query--key channel: correcting only that
route keeps training stable while each fault stays active, and removing the channel's own dominant directions
suppresses the runaway where equal-energy removal elsewhere does not. For the injected error, temporal sign-coherence decides
entry, but ordinary learning is coherent too, so a monitor watches the resulting
attention-logit saturation. QK-Guard acts on that signal, normalizing the query--key projections per head, so
one action contains faults that share no source. Where a source-local repair exists, such as an fp32 streaming
core, it remains the direct fix; the channel is where to act for sources that lack one or keep low precision by
design. Fault source is not
failure channel.

\bibliographystyle{plainnat}
\bibliography{main}

\let\appendixsectionorig\section
\renewcommand{\section}{\FloatBarrier\appendixsectionorig}
\appendix

\providecommand{\maintextname}{main paper}
\providecommand{\paperpartscope}{main text and supplement alike}
\providecommand{\artifactname}{public artifact}

\section{Experimental Setup, Protocols, and Measurement Semantics}
\label{app:setup}

\subsection{Models, data, and optimization}

All main-grid experiments train GPT-2-class decoder-only transformers~\citep{radford-gpt2-2019} on
OpenWebText~\citep{gokaslan-openwebtext-2019} with AdamW~\citep{loshchilov-decoupled-2019}
(\(\beta_1=0.9\), \(\beta_2=0.95\), weight decay \(0\)), learning rate \(10^{-3}\) with 2{,}000
warmup steps and cosine decay toward \(10^{-5}\) on a decay horizon fixed far beyond the length of
every run reported here --- so these runs end while the rate is still near its peak, not at the
floor; the \artifactname{} gives the constant --- gradient clipping at \(1.0\), block size 1{,}024, and
micro-batch size 20 with gradient accumulation set per arm. The base model is GPT-2
small (12 layers, 12 heads, width 768); the variants of the main text modify one component
at a time (RoPE~\citep{su-roformer-2021}, RMSNorm~\citep{zhang-root-2019}, a GPT-2M and a 350M scaling, and a LLaMA-style
architecture~\citep{touvron-llama-2023} combining them). Training is distributed data-parallel;
recorded process counts vary by arm family, including two for the direction-removal battery of \S4.2, four for the
matched-structure assay, and eight for the operand-level arms. We take \texttt{nproc} from the
per-run configurations that record it. The main text states the
shape wherever it bears on a result (the two- versus eight-rank contrast of
Appendix~\ref{app:boundaries}).

The remaining protocol is the nanoGPT recipe, stated here so the grid is reproducible without
reading our code. The corpus is the public OpenWebText release, split into train and validation by
a shuffled split holding out a fraction 0.0005 under split seed 2357 --- both fixed once and shared
by every arm, so all reported validation losses are measured on one held-out set. Text is tokenized
with the GPT-2 byte-pair encoder, documents separated by the
end-of-text token and packed into one flat token stream from which each micro-batch draws
uniformly at random; there is no document masking, so a sequence may span a document boundary.
Gradient accumulation is specified globally and divided across the ranks, and the data sampler is
seeded per run rather than per rank, so every rank draws the same index stream and the ranks
average numerically different copies of one gradient. Two counts therefore differ and we give both.
The tokens \emph{processed} per optimizer step are the global accumulation count times the
micro-batch times the block size; the \emph{distinct} tokens sampled per step are that quantity
divided by the rank count. The global accumulation count is 16 for the main grid and 8 for the
batch-matched 350M pair, so the two-rank direction-removal arms sample half, and the eight-rank
operand-level arms an eighth, of the processed figure. Every load-bearing comparison in this paper
is between arms at a single rank count, so the arms of a contrast always share one data stream and
one distinct-token batch; what the distinction changes is what a reader must hold fixed to
reproduce an arm, and how the two- versus eight-rank contrast of
Appendix~\ref{app:boundaries} should be read. Weights are initialized normal with standard deviation 0.02,
except the residual output projections, which use \(0.02/\sqrt{2L}\) for \(L\) layers as in GPT-2;
the token embedding and the output head are tied. The objective is mean token-level cross-entropy
on next-token prediction. Reported train and validation losses are that same objective averaged
over freshly drawn evaluation batches from the corresponding split, with the model in evaluation
mode, computed at the evaluation interval that also drives the monitor
(Appendix~\ref{app:interventions}); every loss in this paper is this language-modelling loss, in
nats. The evaluation batch is 16 sequences, drawn
independently of the training sampler and independent of the micro-batch size, and the number of
such batches averaged is the per-family evaluation-iteration count; both are fixed within an arm
family, so a reported loss is comparable across the arms of any contrast.

Three random-number streams are seeded separately, and a run is identified by one integer. The
global PyTorch generator, which drives initialization and dropout, is seeded with that integer plus
the rank, so ranks differ where they must. The training data sampler has its own generator seeded
from the run integer alone, without a rank term --- this is what makes the ranks share one index
stream, as described above --- so which indices the stream yields depends only on the run, though
how many of them an optimizer step consumes depends on the rank count, as above. The quantizer's
stochastic-rounding stream is reseeded every step from the step index, so the random variates an
arm draws are a deterministic function of its step; the injected error is those variates applied to
the current gradient.

The main experimental grid runs on NVIDIA H20 GPUs under Linux with torch 2.10 and CUDA 12.8,
distributed over the per-arm rank counts given above. The cross-hardware portability check of \S3.3
of the \maintextname{} runs the same training program that produces the main-grid collapses on a single
NVIDIA B200 (torch 2.11.0+cu128, single GPU, no DDP), with the same software-emulated quantizer: an
exact-code reproduction on a different GPU architecture.

One family departs from the AdamW recipe above. To test whether an adaptive optimizer is required
for the failure, we swap AdamW for SGD and hold the rest of the arm fixed --- same data, same block
size, same injected fault, same crash rule. The learning rates are
\(3{\times}10^{-2}\) and \(10^{-1}\), each with a clean control at the same rate; the crash steps,
horizons and spectral bound in Table~\ref{tab:inventory} are computed from each run's recorded loss
and spectral-norm history by the same extractors used everywhere else, and those records
do not carry the momentum setting or rank count. The family supports the qualitative statement of
\S3.3 of the \maintextname{}: the untreated fault collapses under SGD too, while the matched clean
control at the same learning rate does not.

\subsection{The sustained-divergence crash rule}

Every ``collapse'' verdict in the paper, \paperpartscope{}, is decided by one rule
applied to the logged training loss. Let \(m_t\) be the running minimum of the loss up to step
\(t\). A run's crash step is the first step whose loss exceeds \(m_t + 1.5\) and after which the
loss never again falls below \(m_t + 0.75\). A run with no such step is a survivor and is reported
with the last step it logged; an excursion that recovers is not a crash. The rule is deliberately
insensitive to how long a diverged run kept logging, which also makes post-collapse maxima of any
quantity non-comparable across arms --- the reason doomed traces in the main-text figures stop at
their crash step.

\subsection{Measurement conventions and injection protocols}

\(\sigma(W^Q)\) denotes the spectral norm of a layer's query projection; reported values take the
maximum over layers, and ``max over the run'' additionally takes the maximum over logged steps
--- at the same per-family cadence as the monitor, listed in
Appendix~\ref{app:interventions}.

The attention-logit statistic the monitor reads is an aggregate, not a global maximum, and the
distinction matters for anyone reimplementing the trigger. On a fixed four-example validation probe
batch --- never a training batch, so the measurement cannot perturb the trajectory --- each layer
forms the full causal logit matrix \(A=QK^{\top}/\sqrt{d_h}\) on the same \(Q,K\) the layer just
used, after RoPE where RoPE is active. For that layer we take the row maximum
\(\max_j A_{ij}\), the largest logit each query attends to, and average it over examples, heads,
and query positions; the reported value is then the maximum of that per-layer average over layers.
Averaging inside the layer reduces sensitivity to isolated outliers: a crossing means some layer's
mean row maximum exceeds the threshold. Values elsewhere in the paper labelled ``maximum attention
logit'' are this statistic.

The controlled-error assay of \S2 and \S4 computes, at the instrumented attention backward, both
the faulty and the exact quantity, forms the error, and injects a manipulated version of that error
while the forward pass runs either clean (the clean-forward surrogate) or with the natural
low-precision path enabled (natural context); the two are named per arm and are not
interchangeable. The manipulations used in the paper, each stated by what it does to the injected
tensor: \emph{scalar-mean-only} injects the mean of the current tile; \emph{mean-removed} injects
the error minus that scalar; \emph{per-head DC} variants inject or remove the per-head mean, at
per-tile granularity in the legacy arms and reduced over the whole pass and across ranks in the
full-pass arms (the main text names which is which wherever the difference matters);
\emph{sign-scramble} redraws every element's sign; \emph{batch permutation} permutes the error
along the batch axis, preserving the batch-summed tensor at every head and sequence position;
\emph{sign randomization} assigns random global signs to whole-tensor injection events, or an independent sign per head once per optimizer step;
\emph{matched Gaussian} replaces the error by
noise at the same root-mean-square magnitude. Injection magnitude is scaled by \(\lambda\), with
\(\lambda=1\) the natural size.

The post-reduction weight-gradient quantizer of \S3 and \S4 takes the materialized weight gradient
after backward and reduction and before clipping and the optimizer step, and rounds it onto an
MXFP4-style grid~\citep{rouhani-microscaling-2023} (four-bit elements in blocks of thirty-two
sharing a power-of-two scale); which
module's gradient receives the grid is the arm's one moving part.

\subsection{The interventions, defined}
\label{app:interventions}

Three actions in \S5 fire at the same trigger and fire step, so what separates them is exactly
their definitions.

\emph{Per-head query--key normalization}, the action QK-Guard switches on, root-mean-square
normalizes each query and key vector over its own head dimension, independently per example, head,
and token position:
\(q \mapsto q/\sqrt{\tfrac{1}{d_h}\sum_{c} q_c^{2} + \varepsilon}\) with
\(\varepsilon=10^{-6}\) and the statistic accumulated in fp32, and likewise for \(k\). It
introduces no learnable gain --- this is what ``parameter-free'' means, and it is why a guarded arm
and its unguarded twin have identical parameter counts. It is applied to \(Q\) and \(K\) before the
rotary embedding and before the attention kernel dispatch, so it composes with every attention
implementation and with RoPE rather than replacing either.

\emph{Query--key clipping}, the second query--key-local action, instead bounds the worst case
attainable logit. It uses the Cauchy--Schwarz bound
\(\max_{ij} A_{ij} \le \max_i\lVert q_i\rVert\,\max_j\lVert k_j\rVert/\sqrt{d_h}\) and, when that
bound exceeds the threshold, rescales both \(Q\) and \(K\) by the square root of the ratio, leaving
the bound at the threshold. It is a norm clamp, not a normalization: it acts only when the bound is
exceeded and it preserves the relative geometry within \(Q\) and within \(K\).

\emph{The softmax-local dynamic maximum}, the non-query--key control, changes the shift used by the
streaming softmax rather than the projections. When more than one entry of the current block row
lies within \(10^{-3}\) of that block row's maximum, the shift is doubled where the maximum is
positive and set to zero where it is negative, which is the published remedy for the rounding bias
in this recurrence~\citep{qiu-why-2025}. It therefore acts on how the attention output is accumulated, and leaves \(Q\), \(K\),
and the logits they produce untouched.

The monitor evaluates the statistic of Appendix~A.3 on its fixed probe batch every evaluation
interval, and the interval is a property of the arm family rather than a single constant: 50 steps
for the plain-fault rescue arms, the original 350M triplet, and the RoPE replay and from-scratch
arms; 100 steps for the six matched RoPE pairs, the six further 350M seeds, the held-out
weight-gradient arms, and the 60k durability arms; 200 steps on B200. A crossing is therefore
detected only at an evaluation, which is why
firing values can sit far above the threshold --- the 350M hot-swap of \S5.3 fires at step
\NScaleThreeFiveZeroGptTwomRescFpThreeTwoSZeroFireStep{} on a reading of
\NScaleThreeFiveZeroGptTwomRescFpThreeTwoSZeroFireValue{}, the first reading to cross, having been
below the threshold at the preceding evaluation. The statistic is read
on the post-update weights; once fired, the action is active from the next step and stays on for the
remainder of the run.

\subsection{The threshold of 30: provenance and sweeps}
\label{app:threshold}

The transfer results in \S5.3 use the threshold unchanged; this subsection gives its provenance and
how the outcome responds when it moves.

The value was fixed on the plain GPT-2 accumulator fault when the closed loop was first built,
from the gap between the maximum-logit envelopes that healthy runs reached and the far larger values
doomed runs reached before collapsing. RoPE and 350M runs had been observed by then; every
guarded run reported here was launched afterwards.

The value was not re-tuned: the RoPE, 350M, held-out all-weight-gradient and B200 transfer runs
all used 30. A later sweep also tested
100, the value Kimi~K2 reports at
production scale~\citep{team-kimi-2025}.

\paragraph{What moving the threshold does, at one seed.}
One seed of the held-out all-weight-gradient MXFP4 protocol carries the trigger at three settings
spanning a factor of more than thirty, with the action and everything else held fixed. Raising the
value delays the fire and does not change the outcome within the observed horizons:
\(\tau=30\) fires at step \NEthreeRescSTwoQknThirtyFireStep{} at a monitored logit of
\NEthreeRescSTwoQknThirtyFireLogit{}, \(\tau=100\) at step \NEthreeRescSTwoQknHundredFireStep{}
(\NEthreeRescSTwoQknHundredFireLogit{}), and \(\tau=1{,}000\) at step
\NEthreeRescSTwoQknThousandFireStep{} (\NEthreeRescSTwoQknThousandFireLogit{}) --- the last only
once the runaway has already carried \(\sigma(W^Q)\) to
\NEthreeRescSTwoQknThousandPrefireSigma{}. All three fire before this seed's untreated twin
collapses at \NEthreeMxfpFourSTwoCrash{}, and none of the three collapses inside its own horizon:
the \(\tau=100\) and \(\tau=1{,}000\) arms complete \NEthreeRescSTwoQknThousandStableTo{} logged steps
at \NEthreeRescSTwoQknHundredFinalVal{} and \NEthreeRescSTwoQknThousandFinalVal{} validation loss,
and the \(\tau=30\) arm reaches \NEthreeRescSTwoQknThirtyStableTo{} at \NEthreeRescSTwoQknThirtyFinalVal{}. At \(\tau=100\) the setting repeats across three seeds, firing at steps
\NEthreeRescSZeroQknHundredFireStep{}, \NEthreeRescSOneQknHundredFireStep{} and
\NEthreeRescSTwoQknHundredFireStep{} with none collapsing, and the query--key clip of \S5.2
substitutes for the normalization at that same value in two seeds, at
\NEthreeRescSZeroClipHundredFireStep{} and \NEthreeRescSTwoClipHundredFireStep{}.

On this seed the outcome holds from \(\tau=30\) to \(\tau=1{,}000\), and the two-seed sweep that
follows covers settings from 10 to 100.

\paragraph{Moving the threshold across two seeds.}
Table~\ref{tab:tsweep} runs the unchanged controller at settings from 10 to 100, two seeds each, on the
plain GPT-2 accumulator fault and on the held-out all-weight-gradient MXFP4 fault. A fire is the first
monitored evaluation strictly above the setting, read every 50 steps on the plain fault and every 100 steps
on the held-out fault. On the plain fault every setting fires at the evaluation at step
\NRrOneCalTTenSZeroFireStep{} or at the next one, 50 steps later, and no guarded arm collapses before its
last logged step, while both untreated seeds collapse, at \NRrOneCalNofixSOneCrash{} and
\NRrOneCalNofixSZeroCrash{}. On the held-out fault every setting fires and no arm collapses through
\NRrOneExtTTenSZeroStableTo{} logged steps.

\begin{table}[t]
\centering
\caption{The fixed trigger moved from 10 to 100, two seeds per setting, with the action (per-head QK
normalization) held fixed. \(\blacktriangle\)~first monitored evaluation strictly above the setting, with the
monitored statistic there; \(\times\)~sustained-rule collapse step; \(\blacktriangleright\)~last logged step
without collapse; \(^{c}\)~ended by a four-hour wall-clock cap before the 5,000-step budget. Val.\ loss is the
final validation loss. These launches are separate from those drawn in Figure~\ref{fig:guard}.}
\label{tab:tsweep}
\tabsetup
\begin{tabularx}{\textwidth}{@{\hspace{\tabpad}}Y Q{0.07\textwidth} R{0.08\textwidth} R{0.085\textwidth} P{0.13\textwidth} R{0.11\textwidth}@{\hspace{\tabpad}}}
\toprule
\tabheadrow\colhead{Setting} & \colhead{Seed} & \multicolumn{2}{c}{\colhead{\(\blacktriangle\)~fire (statistic)}} & \colhead{Outcome} & \colhead{Val.\ loss} \\
\midrule
\tabgroup{6}{Plain GPT-2 accumulator fault, cadence 50}\\
10 & 0 & \NRrOneCalTTenSZeroFireStep{} & (\NRrOneCalTTenSZeroFireValue{}) & \(\blacktriangleright\)\,\NRrOneCalTTenSZeroStableTo{}\(^{c}\) & \NRrOneCalTTenSZeroFinalVal{} \\
10 & 1 & \NRrOneCalTTenSOneFireStep{} & (\NRrOneCalTTenSOneFireValue{}) & \(\blacktriangleright\)\,\NRrOneCalTTenSOneStableTo{}\(^{c}\) & \NRrOneCalTTenSOneFinalVal{} \\
20 & 0 & \NRrOneCalTTwentySZeroFireStep{} & (\NRrOneCalTTwentySZeroFireValue{}) & \(\blacktriangleright\)\,\NRrOneCalTTwentySZeroStableTo{}\(^{c}\) & \NRrOneCalTTwentySZeroFinalVal{} \\
20 & 1 & \NRrOneCalTTwentySOneFireStep{} & (\NRrOneCalTTwentySOneFireValue{}) & \(\blacktriangleright\)\,\NRrOneCalTTwentySOneStableTo{}\(^{c}\) & \NRrOneCalTTwentySOneFinalVal{} \\
30 & 0 & \NRrOneCalTThirtySZeroFireStep{} & (\NRrOneCalTThirtySZeroFireValue{}) & \(\blacktriangleright\)\,\NRrOneCalTThirtySZeroStableTo{}\(^{c}\) & \NRrOneCalTThirtySZeroFinalVal{} \\
30 & 1 & \NRrOneCalTThirtySOneFireStep{} & (\NRrOneCalTThirtySOneFireValue{}) & \(\blacktriangleright\)\,\NRrOneCalTThirtySOneStableTo{}\(^{c}\) & \NRrOneCalTThirtySOneFinalVal{} \\
50 & 0 & \NRrOneCalTFiftySZeroFireStep{} & (\NRrOneCalTFiftySZeroFireValue{}) & \(\blacktriangleright\)\,\NRrOneCalTFiftySZeroStableTo{}\(^{c}\) & \NRrOneCalTFiftySZeroFinalVal{} \\
50 & 1 & \NRrOneCalTFiftySOneFireStep{} & (\NRrOneCalTFiftySOneFireValue{}) & \(\blacktriangleright\)\,\NRrOneCalTFiftySOneStableTo{}\(^{c}\) & \NRrOneCalTFiftySOneFinalVal{} \\
100 & 0 & \NRrOneCalTHundredSZeroFireStep{} & (\NRrOneCalTHundredSZeroFireValue{}) & \(\blacktriangleright\)\,\NRrOneCalTHundredSZeroStableTo{}\(^{c}\) & \NRrOneCalTHundredSZeroFinalVal{} \\
100 & 1 & \NRrOneCalTHundredSOneFireStep{} & (\NRrOneCalTHundredSOneFireValue{}) & \(\blacktriangleright\)\,\NRrOneCalTHundredSOneStableTo{}\(^{c}\) & \NRrOneCalTHundredSOneFinalVal{} \\
untreated & 0 & \multicolumn{2}{c}{---} & \(\times\)\,\NRrOneCalNofixSZeroCrash{} & \NRrOneCalNofixSZeroFinalVal{} \\
untreated & 1 & \multicolumn{2}{c}{---} & \(\times\)\,\NRrOneCalNofixSOneCrash{} & \NRrOneCalNofixSOneFinalVal{} \\
\tabgap
\tabgroup{6}{Held-out all-weight-gradient MXFP4, cadence 100}\\
10 & 0 & \NRrOneExtTTenSZeroFireStep{} & (\NRrOneExtTTenSZeroFireValue{}) & \(\blacktriangleright\)\,\NRrOneExtTTenSZeroStableTo{} & \NRrOneExtTTenSZeroFinalVal{} \\
10 & 1 & \NRrOneExtTTenSOneFireStep{} & (\NRrOneExtTTenSOneFireValue{}) & \(\blacktriangleright\)\,\NRrOneExtTTenSOneStableTo{} & \NRrOneExtTTenSOneFinalVal{} \\
20 & 0 & \NRrOneExtTTwentySZeroFireStep{} & (\NRrOneExtTTwentySZeroFireValue{}) & \(\blacktriangleright\)\,\NRrOneExtTTwentySZeroStableTo{} & \NRrOneExtTTwentySZeroFinalVal{} \\
20 & 1 & \NRrOneExtTTwentySOneFireStep{} & (\NRrOneExtTTwentySOneFireValue{}) & \(\blacktriangleright\)\,\NRrOneExtTTwentySOneStableTo{} & \NRrOneExtTTwentySOneFinalVal{} \\
50 & 0 & \NRrOneExtTFiftySZeroFireStep{} & (\NRrOneExtTFiftySZeroFireValue{}) & \(\blacktriangleright\)\,\NRrOneExtTFiftySZeroStableTo{} & \NRrOneExtTFiftySZeroFinalVal{} \\
50 & 1 & \NRrOneExtTFiftySOneFireStep{} & (\NRrOneExtTFiftySOneFireValue{}) & \(\blacktriangleright\)\,\NRrOneExtTFiftySOneStableTo{} & \NRrOneExtTFiftySOneFinalVal{} \\
100 & 0 & \NRrOneExtTHundredSZeroFireStep{} & (\NRrOneExtTHundredSZeroFireValue{}) & \(\blacktriangleright\)\,\NRrOneExtTHundredSZeroStableTo{} & \NRrOneExtTHundredSZeroFinalVal{} \\
100 & 1 & \NRrOneExtTHundredSOneFireStep{} & (\NRrOneExtTHundredSOneFireValue{}) & \(\blacktriangleright\)\,\NRrOneExtTHundredSOneStableTo{} & \NRrOneExtTHundredSOneFinalVal{} \\
\bottomrule
\end{tabularx}
\end{table}

\subsection{Per-arm seed counts and horizons}
\label{app:inventory}

Table~\ref{tab:inventory} lists, for each arm family behind the paper's headline results, the seeds
run, the outcome count, and each survivor's last logged step. Counts are events over the runs
launched; the one LLaMA-architecture launch of the long-horizon sweep whose trajectory matched
neither the collapse signature nor a healthy run is counted in neither class.

A survivor's entry is its last logged step, and a degraded-but-not-collapsed endpoint is recorded
as a survivor with its loss. Horizons, evaluation cadence, architecture, rank count and fault
protocol differ across families, so each count is read within its own family.

\paragraph{Route correction.} The route-correction grid runs in a single process. Each step
backpropagates the same weights and micro-batches through the faulty re-implementation and through the
fp32-core reference, writes the reference gradient into the named route --- for the query--key route,
the query and key rows, weights and biases, of every layer's attention input projection --- and only
then applies global-norm clipping and the AdamW step, so clipping and the optimizer state see the
spliced gradient.

\paragraph{Horizons of the \S3 route-correction grid.} Table~\ref{tab:locus}(a) gives that grid's collapse counts
per cell, each read at its own horizon: on plain GPT-2 the two \(QK\)-corrected arms reach step
\NLocusCorrectGptTwoQkSZeroStableTo{} and the \(V\)/\(O\)-corrected arm was stopped at step
\NLocusCorrectGptTwoVoSZeroStableTo{} (the table's \(\ddagger\)); the 350M \(QK\)-corrected arms reach step
\NLocusCorrectGptTwomQkSZeroStableTo{}, the RoPE arms step \NLocusCorrectRopeQkXOneThreekSZeroStableTo{} and the
LLaMA arms step \NLocusCorrectLlamaQkSZeroStableTo{}, the horizons at which the \(QK\)-corrected finals of \S3.1
are read.

\newcommand{\inventoryfirstrows}{%
\tabgroup{3}{Matched-magnitude structure assay (\S4.1, natural context,
\(\lambda=1\))}\\
true error & 5/5 & $\times$\,\NStructureSFourActualLOneSZeroCrash{}--\NStructureSFourActualLOneSFiveCrash{} \\
sign-scramble & 0/4 & $\blacktriangleright$\,\NStructureSFourScramLOneSZeroStableTo{} (2), \NStructureSFourScramLOneSTwoStableTo{} (2) \\
batch permutation & 3/3 & $\times$\,\NStructureSFourBatchpermLOneSThreeCrash{}--\NStructureSFourBatchpermLOneSFourCrash{} \\
matched Gaussian & 0/2 & $\blacktriangleright$\,\NStructureInjBothRmsnoiseSTwoStableTo{} \\
true error at \(\lambda=0.5\) & 3/3 & $\times$\,\NStructureSFourActualLZeroFiveSTwoCrash{}--\NStructureSFourActualLZeroFiveSOneCrash{} \\
\tabgap
\tabgroup{3}{Mean-component assay (\S4.1)}\\
scalar mean only & 0/2 & $\blacktriangleright$\,\NAnchorBiasonlyBwdSZeroStableTo{}, \NAnchorBiasonlyBwdSOneStableTo{} \\
mean-removed & 2/2 & $\times$\,\NAnchorMeancorrectBwdSOneCrash{}, \NAnchorMeancorrectBwdSZeroCrash{} \\
per-head DC, natural & 6/6 & $\times$\,\NExpSufffullSFiveCrash{}--\NExpSufffullSThreeCrash{} \\
per-head DC, backward-only near threshold & 2/4 & $\times$\,\NAnchorBiasonlyHeadSOneCrash{}, \NAnchorBiasonlyHeadSZeroCrash{}; $\blacktriangleright$\,\NAnchorBiasonlyHeadSThreeStableTo{}, \NAnchorBiasonlyHeadSTwoStableTo{} \\
per-head DC removed, clean forward & 0/2 & $\blacktriangleright$\,\NAnchorMeancorrectHeadSZeroStableTo{} \\
per-head DC removed, natural & 2/2 & $\times$\,\NExpNecSZeroCrash{}, \NExpNecSOneCrash{} \\
\tabgap
\tabgroup{3}{Sign-randomization arms (\S4.1)}\\
whole-tensor sign randomized; grouping not recorded & 0/2 & $\blacktriangleright$\,\NFixSignflipLongSZeroStableTo{} \\
per-head signs per step & 0/2 & $\blacktriangleright$\,\NFixBhSignflipSZeroStableTo{} \\
\tabgap
\tabgroup{3}{Direction-removal battery (\S4.2 and Appendix~\ref{app:battery}; per source
\(\times\) seed)}\\
on-target removal, 4 arms & 0/4 & $\blacktriangleright$\,\NDyadNTwoQkToprenormSZeroStable{} \\
equal-energy off-target, model-wide & 2/2 & $\times$\,\NDyadNTwoAllOfftgtrenormTwoSZeroCrash{}, \NDyadNTwoAllOfftgtrenormTwoSOneCrash{} \\
\;\;query--key & 0/2 & $\blacktriangleright$\,\NDyadNTwoQkOfftgtrenormTwoSZeroNoCrashTo{} \\
\;\;query--key, 32k schedule & \NDyadNTwoQkOfftgtThreeTwoKCrashedCount{}/4 & $\times$\,\NDyadNTwoQkOfftgtThreeTwoKSTwoCrash{}--\NDyadNTwoQkOfftgtThreeTwoKSThreeCrash{} \\
isotropic shrink, 4 arms & 4/4 & $\times$\,\NDyadNTwoQkShrinkSZeroCrash{}--\NDyadNTwoAllShrinkSZeroCrash{} \\
random-\(k\) removal, 4 arms & 4/4 & $\times$\,\NDyadNtwoAllRandkSOneCrash{}--\NDyadNtwoAllRandkSZeroCrash{} \\
350M on-target & 0/2 & $\blacktriangleright$\,\NDyadThreeFiftyQkToprenormSZeroStable{} \\
\;\;350M off-target & 2/2 & $\times$\,\NDyadThreeFiftyQkOfftgtrenormSOneCrash{}, \NDyadThreeFiftyQkOfftgtrenormSZeroCrash{} \\
}
\newcommand{\inventorysecondrows}{%
\tabgroup{3}{Closed-loop transfer replication (\S5.3)}\\
RoPE from-scratch pairs, guarded & 0/6 & $\blacktriangleright$\,\NRopeRopeThreeCtrlSOneStableTo{} \\
\;\;their unguarded twins & 6/6 & $\times$\,\NRopeRopeThreeNorescSThreeCrash{}--\NRopeRopeThreeNorescSOneCrash{} \\
350M at $2\times$ batch, unguarded & 4/4 & $\times$\,\NScaleThreeFiveZeroGThreeFiveZeroRescNoneSZeroCrash{}--\NScaleThreeFiveZeroGThreeFiveZeroRescNoneSThreeCrash{} \\
\;\;its rescued arms & 0/8 & $\blacktriangleright$\,\NScaleThreeFiveZeroGThreeFiveZeroRescQknSZeroStableTo{} \\
350M batch-matched, unguarded & 2/2 & $\times$\,\NScaleThreeFiveZeroGThreeFiveZerobRescNoneSOneCrash{}, \NScaleThreeFiveZeroGThreeFiveZerobRescNoneSZeroCrash{} \\
\;\;its rescued arms & 0/4 & $\blacktriangleright$\,\NScaleThreeFiveZeroGThreeFiveZerobRescQknSZeroStableTo{} \\
\tabgap
\tabgroup{3}{Deployment durability (\S5.4; 3 seeds per method)}\\
unprotected & 3/3 & $\times$\,\NHorizonLhNofixSZeroCrash{}, \NHorizonLhNofixSTwoCrash{}, \NHorizonLhNofixSOneCrash{} \\
always-on QK-norm & 0/3 & $\blacktriangleright$\,\NHorizonLhQknStableTo{} \\
QK-Guard & 0/3 & $\blacktriangleright$\,\NHorizonLhCtrlStableTo{} \\
\tabgap
\tabgroup{3}{Optimizer generality (\S3.3; SGD in place of AdamW)}\\
untreated fault, LR \(3{\times}10^{-2}\) & 2/2 & $\times$\,\NOptSgdDirtyLrThreeemTwoSOneCrash{}, \NOptSgdDirtyLrThreeemTwoSZeroCrash{} \\
\;\;its matched clean controls & 0/2 & $\blacktriangleright$\,\NOptSgdCleanLrThreeemTwoSZeroStableTo{} (2); final train \NOptSgdCleanLrThreeemTwoSZeroFinal{} \\
untreated fault, LR \(10^{-1}\) & 1/1 & $\times$\,\NOptSgdDirtyLrOneemOneSZeroCrash{}; peak \(\sigma(W^Q)\) \NTriggerSgdDirtyLrOneemOneSZeroMaxSigma{} \\
\;\;its matched clean control & 0/1 & $\blacktriangleright$\,\NOptSgdCleanLrOneemOneSZeroStableTo{}; final train \NOptSgdCleanLrOneemOneSZeroFinal{} \\
\tabgap
\tabgroup{3}{Source shift at a further seed (Figure~\ref{fig:dissociation}'s arms, seed~2)}\\
weight-gradient quantizer on \(QK\) & 1/1 & $\times$\,\NRrTwoQkSTwoCrash{} \\
\;\;per-head \(QK\)-norm twin & 0/1 & $\blacktriangleright$\,\NRrTwoQkqknSTwoStableTo{} \\
on MLP & 1/1 & $\times$\,\NRrTwoMlpSTwoCrash{} \\
\;\;per-head \(QK\)-norm twin & 0/1 & $\blacktriangleright$\,\NRrTwoMlpqknSTwoStableTo{} \\
on \(V\) & 1/1 & $\times$\,\NRrTwoVSTwoCrash{} \\
\;\;per-head \(QK\)-norm twin & 0/1 & $\blacktriangleright$\,\NRrTwoVqknSTwoStableTo{} \\
\tabgap
\tabgroup{3}{Boundary arms (Appendix~\ref{app:boundaries})}\\
operand-level GEMMs, fp32 accum & 0/2 & $\blacktriangleright$\,\NEToENofixStableTo{}; registered seed's \(\sigma \le \NEToENofixMaxSigma{}\) \\

}
\ifdefined\arxivlayout
\begin{table*}[p]
\centering
\tabsetupdense
\caption{Seed counts, outcomes, and horizons for the arm families behind the paper's headline
results and the generality checks around them. ``Coll.'' counts collapses over the runs launched;
\(\times\)~marks collapse at the listed step(s) (a range spans earliest--latest);
\(\blacktriangleright\)~a survivor's last logged step, with (2) marking two seeds at that
horizon. Where a row turns on a further quantity --- a spectral bound, a final train or
validation loss --- that quantity is named in place. The optimizer-generality rows use
SGD at their own learning rates (Appendix~\ref{app:setup}). The \S3 route-correction and
source-shift grids report their per-cell seed counts directly in Table~2 of the \maintextname{}; the
statistics-carry route grid reports its two seeds inline in \S3.1.}
\label{tab:inventory}
\begin{tabularx}{\textwidth}{@{\hspace{\tabpad}}P{\tabstubw} Q{\tabcollw} Y@{\hspace{\tabpad}}}
\toprule
\tabheadrow\colhead{Arm family} & \colhead{Coll.} & \colhead{Outcome} \\
\midrule
\inventoryfirstrows
\bottomrule
\end{tabularx}
\end{table*}

\begin{table*}[p]
\ContinuedFloat
\centering
\tabsetupdense
\caption{Seed counts, outcomes, and horizons (continued).}
\begin{tabularx}{\textwidth}{@{\hspace{\tabpad}}P{\tabstubw} Q{\tabcollw} Y@{\hspace{\tabpad}}}
\toprule
\tabheadrow\colhead{Arm family} & \colhead{Coll.} & \colhead{Outcome} \\
\midrule
\inventorysecondrows
\bottomrule
\end{tabularx}
\end{table*}
\else
\begin{table*}[t]
\centering
\footnotesize
\setlength{\tabcolsep}{4pt}
\begin{tabular}{@{}l c l@{}}
\toprule
\tabheadrow\colhead{Arm family} & \colhead{Coll.} & \colhead{Outcome} \\
\midrule
\inventoryfirstrows
\tabgap
\inventorysecondrows
\bottomrule
\end{tabular}
\caption{Seed counts, outcomes, and horizons for the arm families behind the paper's headline
results and the generality checks around them. ``Coll.'' counts collapses over the runs launched;
\(\times\)~marks collapse at the listed step(s) (a range spans earliest--latest);
\(\blacktriangleright\)~a survivor's last logged step, with (2) marking two seeds at that
horizon. Where a row turns on a further quantity --- a spectral bound, a final train or
validation loss --- that quantity is named in place. The optimizer-generality rows use
SGD at their own learning rates (Appendix~\ref{app:setup}). The \S3 route-correction and
source-shift grids report their per-cell seed counts directly in Table~2 of the \maintextname{}; the
statistics-carry route grid reports its two seeds inline in \S3.1.}
\label{tab:inventory}
\end{table*}
\fi

\section{The QK-Subspace Control Battery}
\label{app:battery}

The causal claim of \S4.2 is carried by one contrast --- remove the update's component along the
current weight's three leading query--key singular directions at restored norm, versus remove equal
energy from outside that subspace at matched norm and magnitude. This appendix reports the full
battery around that contrast, all at the GPT-2-small scale on the post-reduction weight-gradient sources (model-wide
and query--key-only), two seeds each, 16k schedules.
Figure~2 of the \maintextname{} shows the primary contrast itself; the arms below surround it.

\paragraph{The two operations.} At each optimizer step, and separately for every layer's \(W^Q\) and
\(W^K\), let \(u_k,v_k\) be the leading singular pairs of the weight before the step and \(\Delta W\) the
update the optimizer applied. On-target removal subtracts
\(q=\sum_k\langle\Delta W,u_kv_k^\top\rangle_F\,u_kv_k^\top\) over the three leading pairs and rescales the
remainder to \(\lVert\Delta W\rVert_F\). The off-target control instead subtracts a component \(s\) built
from the remainder \(\Delta W-q\) and a random unit direction orthogonal to every \(u_kv_k^\top\) and to
that remainder, drawn with the same per-step seed on every rank, such that
\(\lVert s\rVert_F=\lVert q\rVert_F\) and \(\lVert\Delta W-s\rVert_F=\lVert\Delta W-q\rVert_F\), and rescales the
same way; \(s\) has no component along any \(u_kv_k^\top\), so the target component stays in place.
On-target removal fixes only the first-order change of the three current leading singular values: the
rescaling returns the removed energy to the other directions, and \(\sigma(W^Q)\), the largest singular
value over all directions, would register a runaway carried by any of them.

Plain top-three removal without renormalization behaves like the renormalized form: no arm
collapses, three running to \NDyadNtwoAllTopkSZeroStable{} steps and one to
\NDyadNtwoQkTopkSOneStable{}, with \(\sigma(W^Q)\) between \NDyadNtwoQkTopkSZeroMaxSigma{} and
\NDyadNtwoAllTopkSOneMaxSigma{}, so the renormalization is not what buys survival. Projecting the
update off three \emph{random} rank-one directions instead --- matched in rank, not in removed
energy, since a random direction carries almost none of the update's mass --- does not suppress:
all four arms collapse, between \NDyadNtwoAllRandkSOneCrash{} and \NDyadNtwoAllRandkSZeroCrash{}
steps, with \(\sigma(W^Q)\) reaching \NDyadNtwoAllRandkSZeroMaxSigma{} to
\NDyadNtwoQkRandkSZeroMaxSigma{}.
Shrinking the whole update isotropically by exactly the energy the top-three removal takes --- the
pure dose control --- also does not suppress: all four arms collapse between
\NDyadNTwoQkShrinkSZeroCrash{} and \NDyadNTwoAllShrinkSZeroCrash{} steps. The equal-energy
off-target arms and the 32k re-runs are reported in the main text and Table~\ref{tab:inventory};
together the battery isolates \emph{direction}, not update-energy dose, as what the suppression
tracks.

\section{Full Per-Arm Results}
\label{app:tables}

This appendix collects per-arm values that the main text summarizes at the outcome level.
Table~\ref{tab:guard} lists what the closed loop of \S5 was tested against, grouped by the
question each contrast answers, and Table~\ref{tab:durability} reports the deployment-durability
endpoints behind \S5.4: three seeds per method over a 60k-step schedule, where the survivors'
final validation losses agree to within a few hundredths of a nat while every unprotected control
collapses in the first quarter of the schedule.

Table~\ref{tab:guard}'s final group runs the guard on training with no fault at all. Its three arms
share seed, world size, schedule and source files, and differ only in what is switched on ---
nothing, the guard, or always-on \(QK\)-normalisation. The unguarded twin crosses the fixed
threshold of 30 without collapsing, so the guard fires on its counterpart at
\NAccTOneSdpaCtrlSOneOneZerokFireStep{}, and the armed run ends within a few hundredths of a nat of
the always-on arm.

\paragraph{Same trigger, same nominal bound, inside the softmax.}
Section~5.2's same-trigger comparison sets a query--key clip against a softmax-local action that bounds
nothing. Table~\ref{tab:actlocus} adds softmax-local actions that do bound the scores: after firing,
every attention score \(s\) inside the streaming softmax is replaced by \(c\tanh(s/c)\), with \(c=30\),
the clip's nominal bound, or \(c=8\), per-head QK normalization's nominal bound, and with the trigger,
threshold and fire step unchanged. The cap matches only that nominal bound, not the distribution of the
logits the softmax sees, and its derivative also nearly removes the gradient that saturated scores send
back to the query and key. Classes were fixed before the runs, against the seed-matched QK-normalized
arm at each seed's common last logged step \(H\): restored with no sustained collapse through \(H\) and a
validation-loss gap within \(+0.15\) nats, failed on a sustained collapse or a gap of \(+0.5\) or more,
partial in between; no cell was partial. The cap at 30 restores all four seeds, as the clip does, and so
does the cap at 8. A straight-through variant, which keeps the cap's forward values and omits only the
tanh derivative factor, collapses on all four seeds, and the dynamic maximum, rerun here to the 5,000-step
budget, collapses on all four as well; Table~\ref{tab:guard} keeps the endpoints of its shorter earlier runs.
At this nominal bound, then, a softmax-local transformation rescues as the query--key-local
actions do, and removing the tanh derivative attenuation prevented rescue on these seeds.

\begin{table}[t]
\centering
\caption{The same trigger (threshold 30) followed by different actions, four seeds each. \(\blacktriangle\)~first
monitored evaluation above the threshold, with the statistic there; validation loss and its gap to the seed's
QK-normalized reference at the seed's common horizon \(H\) (restored within \(+0.15\) nats and without a
sustained collapse); full run: \(\times\)~sustained-rule collapse step, \(\blacktriangleright\)~last logged step
without collapse. Seeds 0--1 reuse the trigger sweep's QK-normalized and untreated arms, whose wall-clock cap sets \(H\)
there.}
\label{tab:actlocus}
\tabsetup
\begin{tabularx}{\textwidth}{@{\hspace{\tabpad}}Y R{0.08\textwidth} R{0.085\textwidth} R{0.12\textwidth} R{0.13\textwidth} P{0.12\textwidth}@{\hspace{\tabpad}}}
\toprule
\tabheadrow\colhead{Action after the fire} & \multicolumn{2}{c}{\colhead{\(\blacktriangle\)~fire (statistic)}} & \colhead{val.\ at \(H\)} & \colhead{gap at \(H\)} & \colhead{full run} \\
\midrule
\tabgroup{6}{seed~0, common horizon \(H=\)\,\NActHPrimarySZero{}}\\
QK-normalized reference & \NRrOneCalTThirtySZeroFireStep{} & (\NRrOneCalTThirtySZeroFireValue{}) & \NActScapThirtySZeroValRefH{} & \multicolumn{1}{c}{---} & \(\blacktriangleright\)\,\NRrOneCalTThirtySZeroStableTo{} \\
no action & \multicolumn{2}{c}{---} & \multicolumn{1}{c}{---} & \multicolumn{1}{c}{---} & \(\times\)\,\NRrOneCalNofixSZeroCrash{} \\
QK clip, bound 30 & \NActQkclipSZeroFireStep{} & (\NActQkclipSZeroFireValue{}) & \NActQkclipSZeroValH{} & \tabsign{\NActQkclipSZeroGapH{}} & \(\blacktriangleright\)\,\NActQkclipSZeroStableTo{} \\
tanh cap, \(c=30\) & \NActScapThirtySZeroFireStep{} & (\NActScapThirtySZeroFireValue{}) & \NActScapThirtySZeroValH{} & \tabsign{\NActScapThirtySZeroGapH{}} & \(\blacktriangleright\)\,\NActScapThirtySZeroStableTo{} \\
tanh cap, \(c=8\) & \NActScapEightSZeroFireStep{} & (\NActScapEightSZeroFireValue{}) & \NActScapEightSZeroValH{} & \tabsign{\NActScapEightSZeroGapH{}} & \(\blacktriangleright\)\,\NActScapEightSZeroStableTo{} \\
\;\;straight-through, \(c=30\) & \NActScapsteThirtySZeroFireStep{} & (\NActScapsteThirtySZeroFireValue{}) & \NActScapsteThirtySZeroValH{} & \tabsign{\NActScapsteThirtySZeroGapH{}} & \(\times\)\,\NActScapsteThirtySZeroCrash{} \\
dynamic maximum & \NActDynmaxSZeroFireStep{} & (\NActDynmaxSZeroFireValue{}) & \NActDynmaxSZeroValH{} & \tabsign{\NActDynmaxSZeroGapH{}} & \(\times\)\,\NActDynmaxSZeroCrash{} \\
\tabgap
\tabgroup{6}{seed~1, common horizon \(H=\)\,\NActHPrimarySOne{}}\\
QK-normalized reference & \NRrOneCalTThirtySOneFireStep{} & (\NRrOneCalTThirtySOneFireValue{}) & \NActScapThirtySOneValRefH{} & \multicolumn{1}{c}{---} & \(\blacktriangleright\)\,\NRrOneCalTThirtySOneStableTo{} \\
no action & \multicolumn{2}{c}{---} & \multicolumn{1}{c}{---} & \multicolumn{1}{c}{---} & \(\times\)\,\NRrOneCalNofixSOneCrash{} \\
QK clip, bound 30 & \NActQkclipSOneFireStep{} & (\NActQkclipSOneFireValue{}) & \NActQkclipSOneValH{} & \tabsign{\NActQkclipSOneGapH{}} & \(\blacktriangleright\)\,\NActQkclipSOneStableTo{} \\
tanh cap, \(c=30\) & \NActScapThirtySOneFireStep{} & (\NActScapThirtySOneFireValue{}) & \NActScapThirtySOneValH{} & \tabsign{\NActScapThirtySOneGapH{}} & \(\blacktriangleright\)\,\NActScapThirtySOneStableTo{} \\
tanh cap, \(c=8\) & \NActScapEightSOneFireStep{} & (\NActScapEightSOneFireValue{}) & \NActScapEightSOneValH{} & \tabsign{\NActScapEightSOneGapH{}} & \(\blacktriangleright\)\,\NActScapEightSOneStableTo{} \\
\;\;straight-through, \(c=30\) & \NActScapsteThirtySOneFireStep{} & (\NActScapsteThirtySOneFireValue{}) & \NActScapsteThirtySOneValH{} & \tabsign{\NActScapsteThirtySOneGapH{}} & \(\times\)\,\NActScapsteThirtySOneCrash{} \\
dynamic maximum & \NActDynmaxSOneFireStep{} & (\NActDynmaxSOneFireValue{}) & \NActDynmaxSOneValH{} & \tabsign{\NActDynmaxSOneGapH{}} & \(\times\)\,\NActDynmaxSOneCrash{} \\
\tabgap
\tabgroup{6}{seed~2, common horizon \(H=\)\,\NActHPrimarySTwo{}}\\
QK-normalized reference & \NActQknThirtySTwoFireStep{} & (\NActQknThirtySTwoFireValue{}) & \NActScapThirtySTwoValRefH{} & \multicolumn{1}{c}{---} & \(\blacktriangleright\)\,\NActQknThirtySTwoStableTo{} \\
no action & \multicolumn{2}{c}{---} & \multicolumn{1}{c}{---} & \multicolumn{1}{c}{---} & \(\times\)\,\NActNofixSTwoCrash{} \\
QK clip, bound 30 & \NActQkclipSTwoFireStep{} & (\NActQkclipSTwoFireValue{}) & \NActQkclipSTwoValH{} & \tabsign{\NActQkclipSTwoGapH{}} & \(\blacktriangleright\)\,\NActQkclipSTwoStableTo{} \\
tanh cap, \(c=30\) & \NActScapThirtySTwoFireStep{} & (\NActScapThirtySTwoFireValue{}) & \NActScapThirtySTwoValH{} & \tabsign{\NActScapThirtySTwoGapH{}} & \(\blacktriangleright\)\,\NActScapThirtySTwoStableTo{} \\
tanh cap, \(c=8\) & \NActScapEightSTwoFireStep{} & (\NActScapEightSTwoFireValue{}) & \NActScapEightSTwoValH{} & \tabsign{\NActScapEightSTwoGapH{}} & \(\blacktriangleright\)\,\NActScapEightSTwoStableTo{} \\
\;\;straight-through, \(c=30\) & \NActScapsteThirtySTwoFireStep{} & (\NActScapsteThirtySTwoFireValue{}) & \NActScapsteThirtySTwoValH{} & \tabsign{\NActScapsteThirtySTwoGapH{}} & \(\times\)\,\NActScapsteThirtySTwoCrash{} \\
dynamic maximum & \NActDynmaxSTwoFireStep{} & (\NActDynmaxSTwoFireValue{}) & \NActDynmaxSTwoValH{} & \tabsign{\NActDynmaxSTwoGapH{}} & \(\times\)\,\NActDynmaxSTwoCrash{} \\
\tabgap
\tabgroup{6}{seed~3, common horizon \(H=\)\,\NActHPrimarySThree{}}\\
QK-normalized reference & \NActQknThirtySThreeFireStep{} & (\NActQknThirtySThreeFireValue{}) & \NActScapThirtySThreeValRefH{} & \multicolumn{1}{c}{---} & \(\blacktriangleright\)\,\NActQknThirtySThreeStableTo{} \\
no action & \multicolumn{2}{c}{---} & \multicolumn{1}{c}{---} & \multicolumn{1}{c}{---} & \(\times\)\,\NActNofixSThreeCrash{} \\
QK clip, bound 30 & \NActQkclipSThreeFireStep{} & (\NActQkclipSThreeFireValue{}) & \NActQkclipSThreeValH{} & \tabsign{\NActQkclipSThreeGapH{}} & \(\blacktriangleright\)\,\NActQkclipSThreeStableTo{} \\
tanh cap, \(c=30\) & \NActScapThirtySThreeFireStep{} & (\NActScapThirtySThreeFireValue{}) & \NActScapThirtySThreeValH{} & \tabsign{\NActScapThirtySThreeGapH{}} & \(\blacktriangleright\)\,\NActScapThirtySThreeStableTo{} \\
tanh cap, \(c=8\) & \NActScapEightSThreeFireStep{} & (\NActScapEightSThreeFireValue{}) & \NActScapEightSThreeValH{} & \tabsign{\NActScapEightSThreeGapH{}} & \(\blacktriangleright\)\,\NActScapEightSThreeStableTo{} \\
\;\;straight-through, \(c=30\) & \NActScapsteThirtySThreeFireStep{} & (\NActScapsteThirtySThreeFireValue{}) & \NActScapsteThirtySThreeValH{} & \tabsign{\NActScapsteThirtySThreeGapH{}} & \(\times\)\,\NActScapsteThirtySThreeCrash{} \\
dynamic maximum & \NActDynmaxSThreeFireStep{} & (\NActDynmaxSThreeFireValue{}) & \NActDynmaxSThreeValH{} & \tabsign{\NActDynmaxSThreeGapH{}} & \(\times\)\,\NActDynmaxSThreeCrash{} \\
\bottomrule
\end{tabularx}
\end{table}

\begin{table*}[t]
\centering
\caption{What the closed loop was tested against, grouped by the question each contrast answers.
``Coll.'' counts runs that ended in a sustained collapse over runs launched; the result cell
gives whichever quantity that row's question turns on and names it ---
\(\blacktriangle\)~fire step, \(\times\)~collapse step, \(\blacktriangleright\)~last logged
step without collapse, ``loss'' a final train loss, ``gap'' a final-loss gap against the
always-on reference. \(^{\dagger}\)The non-collapsing dynamic-max seed plateaus in a degraded state;
neither seed returns to healthy training. \(^{\ddagger}\)Two launches, neither collapsed; the first,
stopped externally before the horizon, is counted as a censored run from its run record, so the
endpoint reported is the second's.}
\label{tab:guard}
\ifdefined\arxivlayout
\tabsetup
\begin{tabularx}{\textwidth}{@{\hspace{\tabpad}}P{\tabstubw} Q{\tabcollw} Y@{\hspace{\tabpad}}}
\else
\footnotesize
\setlength{\tabcolsep}{6pt}
\begin{tabular}{@{}l c l@{}}
\fi
\toprule
\tabheadrow\colhead{Arm} & \colhead{Coll.} & \colhead{Result} \\
\midrule
\tabgroup{3}{Does the alarm fire before the collapse?}\\
plain, unguarded
  & 2/2 & loss \NProtocolRescNoneSZeroFinal{}, \NProtocolRescNoneSOneFinal{} \\
plain, guarded
  & 0/6 & all fire pre-collapse \\
RoPE resume, monitor on
  & 0/1 & loss \NRopeRopeResumeMonSZeroFinal{} \\
\;\;monitor off (same trajectory)
  & 1/1 & loss \NRopeRopeResumeNoneSZeroFinal{} \\
RoPE from scratch, monitored
  & 0/1 & $\blacktriangle$\,\NRopeRopeFullCtrlSZeroBFireStep{} \\
\;\;monitor off (matched config)
  & 1/1 & $\times$\,\NRopeRopeFullNorescSZeroCrash{} \\
\;\;six independent pairs, guarded
  & 0/6 & $\blacktriangle$\,\NRopeRopeThreeCtrlSFourFireStep{}--\NRopeRopeThreeCtrlSOneFireStep{}; $\blacktriangleright$\,\NRopeRopeThreeCtrlSOneStableTo{} \\
\;\;\;\;their unguarded twins
  & 6/6 & $\times$\,\NRopeRopeThreeNorescSThreeCrash{}--\NRopeRopeThreeNorescSOneCrash{} \\
350M, unguarded
  & 1/1 & loss \NScaleThreeFiveZeroGptTwomRescNoneSZeroFinal{} \\
\;\;$QK$-norm action
  & 0/1 & loss \NScaleThreeFiveZeroGptTwomRescLogitSZeroFinal{} \\
\;\;fp32 streaming-core action
  & 0/1 & loss \NScaleThreeFiveZeroGptTwomRescFpThreeTwoSZeroFinal{} \\
\;\;replication at $2\times$ batch, unguarded
  & 4/4 & $\times$\,\NScaleThreeFiveZeroGThreeFiveZeroRescNoneSZeroCrash{}--\NScaleThreeFiveZeroGThreeFiveZeroRescNoneSThreeCrash{} \\
\;\;\;\;its rescued arms, both actions
  & 0/8 & all $\blacktriangle$\,\NScaleThreeFiveZeroGThreeFiveZeroRescQknSZeroFireStep{}; $\blacktriangleright$\,\NScaleThreeFiveZeroGThreeFiveZeroRescQknSZeroStableTo{} \\
\;\;batch-matched replication, unguarded
  & 2/2 & $\times$\,\NScaleThreeFiveZeroGThreeFiveZerobRescNoneSOneCrash{}, \NScaleThreeFiveZeroGThreeFiveZerobRescNoneSZeroCrash{} \\
\;\;\;\;its rescued arms, both actions
  & 0/4 & all $\blacktriangle$\,\NScaleThreeFiveZeroGThreeFiveZerobRescQknSZeroFireStep{}, the recorded fire step \\
\tabgap
\tabgroup{3}{Same trigger, same fire step: does the action matter?}\\
$QK$-local clip
  & 0/2 & loss \NProtocolRescQkclipSZeroFinal{}, \NProtocolRescQkclipSOneFinal{} \\
softmax-local dynamic max
  & 1/2\rlap{$^{\dagger}$} & loss \NProtocolRescDynmaxSZeroFinal{}, \NProtocolRescDynmaxSOneFinal{} \\
\tabgap
\tabgroup{3}{Does it carry to a source the controller never saw?}\\
all-Wgrad MXFP4, QK-Guard
  & 0/2 & $\blacktriangle$\,\NCtrlAllCtrlSZeroFireStep{}, \NCtrlAllCtrlSOneFireStep{}; gap
  \NCtrlGapSZeroFinalVal{}, \NCtrlGapSOneFinalVal{} \\
\;\;always-on $QK$-norm
  & 0/2 & the reference those gaps are taken against \\
\tabgap
\tabgroup{3}{What does arming the guard cost when there is no fault?}\\
unguarded twin
  & 0/1 & $\blacktriangleright$\,\NAccTOneSdpaSOneOneZerokStableTo{}; loss
  \NAccTOneSdpaSOneOneZerokFinal{}, val \NAccTOneSdpaSOneOneZerokFinalVal{} \\
\;\;QK-Guard armed
  & 0/1 & $\blacktriangle$\,\NAccTOneSdpaCtrlSOneOneZerokFireStep{}; loss
  \NAccTOneSdpaCtrlSOneOneZerokFinal{}, val \NAccTOneSdpaCtrlSOneOneZerokFinalVal{} \\
\;\;always-on $QK$-norm
  & 0/2\rlap{$^{\ddagger}$} & loss \NAccTOneSdpaQknSOnebOneZerokFinal{}, val
  \NAccTOneSdpaQknSOnebOneZerokFinalVal{} \\
\bottomrule
\ifdefined\arxivlayout
\end{tabularx}
\else
\end{tabular}
\fi
\end{table*}

\begin{table*}[!t]
\centering
\caption{Deployment durability at the 60k horizon (\S5.4), per seed. For the unprotected
method the cell gives the three crash steps (\(\times\)) under the sustained rule of
Appendix~\ref{app:setup}; for the survivors it gives the shared horizon
(\(\blacktriangleright\), all three seeds) followed by the three final validation losses.
The survivors' peak \(\sigma(W^Q)\) over the horizon reaches
\NHorizonLhSurvivorPeakSigma{}; durability is read from the loss.}
\label{tab:durability}
\ifdefined\arxivlayout
\tabsetup
\begin{tabularx}{\textwidth}{@{\hspace{\tabpad}}P{\tabstubw} Q{\tabcollw} Y@{\hspace{\tabpad}}}
\else
\footnotesize
\setlength{\tabcolsep}{4pt}
\begin{tabular}{@{}l c l@{}}
\fi
\toprule
\tabheadrow\colhead{Method} & \colhead{Coll.} & \colhead{Per-seed outcome} \\
\midrule
none & 3/3 & $\times$\,\NHorizonLhNofixSZeroCrash{}, \NHorizonLhNofixSTwoCrash{}, \NHorizonLhNofixSOneCrash{} \\
QK-norm, always on & 0/3 & $\blacktriangleright$\,\NHorizonLhQknStableTo{} (3); \NHorizonLhQknFinal{}, \NHorizonLhQknSOneFinal{}, \NHorizonLhQknSTwoFinal{} \\
QK-Guard & 0/3 & $\blacktriangleright$\,\NHorizonLhCtrlStableTo{} (3); \NHorizonLhCtrlFinal{}, \NHorizonLhCtrlSOneFinal{}, \NHorizonLhCtrlSTwoFinal{} \\
\bottomrule
\ifdefined\arxivlayout
\end{tabularx}
\else
\end{tabular}
\fi
\end{table*}

\subsection{Positioning and further details}
\begin{table}[!t]
\centering
\caption{What the injected error's structure decides. Rows name the treatment applied, not a property of the
resulting trajectory. Panel (a) holds per-injection magnitude fixed, (b) varies the component, (c) compares whole-tensor
sign randomization at unchanged per-injection magnitude with per-head sign randomization at exact per-step
carrier magnitude. ``Coll.'' gives collapsed runs over runs tested; \(\times\) the collapse step(s),
\(\blacktriangleright\) the last logged step without collapse, (2)~two seeds sharing that horizon,
``bwd.''~backward.}
\label{tab:coherence}
\ifdefined\arxivlayout
\tabsetup
\begin{tabularx}{\textwidth}{@{\hspace{\tabpad}}P{\tabstubw} Q{\tabcollw} Y@{\hspace{\tabpad}}}
\else
\footnotesize
\setlength{\tabcolsep}{3pt}
\begin{tabular}{@{}l c >{\raggedright\arraybackslash}p{0.36\columnwidth}@{}}
\fi
\toprule
\tabheadrow\colhead{Injected-error treatment} & \colhead{Coll.} & \colhead{Step} \\
\midrule
\tabpanel{3}{(a)~Destroy one property, at matched magnitude}\\
nothing (the true error) & 5/5
  & $\times$\,\NStructureSFourActualLOneSZeroCrash--\NStructureSFourActualLOneSFiveCrash{} \\
permuted across the batch & 3/3
  & $\times$\,\NStructureSFourBatchpermLOneSThreeCrash--\NStructureSFourBatchpermLOneSFourCrash{} \\
sign scrambled per element & 0/4
  & $\blacktriangleright$\,\NStructureSFourScramLOneSZeroStableTo{} (2), \NStructureSFourScramLOneSTwoStableTo{} (2) \\
replaced by Gaussian noise & 0/2
  & $\blacktriangleright$\,\NStructureInjBothRmsnoiseSTwoStableTo{} \\
\tabgap
\tabpanel{3}{(b)~Inject or remove one component}\\
\tabgroup{3}{injected:}\\
\;\;per-tile scalar mean only & 0/2
  & $\blacktriangleright$\,\NAnchorBiasonlyBwdSZeroStableTo{}, \NAnchorBiasonlyBwdSOneStableTo{} \\
\;\;the error minus that scalar & 2/2
  & $\times$\,\NAnchorMeancorrectBwdSOneCrash{}, \NAnchorMeancorrectBwdSZeroCrash{} \\
\;\;full-pass per-head DC, natural & 6/6
  & $\times$\,\NExpSufffullSFiveCrash--\NExpSufffullSThreeCrash{} \\
\;\;per-tile per-head DC, bwd. & 2/4
  & $\times$\,\NAnchorBiasonlyHeadSOneCrash{}, \NAnchorBiasonlyHeadSZeroCrash{}; $\blacktriangleright$\,\NAnchorBiasonlyHeadSThreeStableTo{}, \NAnchorBiasonlyHeadSTwoStableTo{} \\
\tabgap
\tabgroup{3}{removed:}\\
\;\;per-tile per-head DC, clean fwd & 0/2
  & $\blacktriangleright$\,\NAnchorMeancorrectHeadSZeroStableTo{} \\
\;\;full-pass per-head DC, natural & 2/2
  & $\times$\,\NExpNecSZeroCrash{}, \NExpNecSOneCrash{} \\
\tabgap
\tabpanel{3}{(c)~Randomize its sign, at matched magnitude}\\
global sign randomized; grouping not recorded & 0/2
  & $\blacktriangleright$\,\NFixSignflipLongSZeroStableTo{} \\
one sign per head, per step & 0/2
  & $\blacktriangleright$\,\NFixBhSignflipSZeroStableTo{} \\
\bottomrule
\ifdefined\arxivlayout
\end{tabularx}
\else
\end{tabular}
\fi
\end{table}

\paragraph{Related interventions.} Static query--key normalization starts at step
zero~\citep{henry-querykey-2020,dehghani-scaling-2023,rybakov-methods-2024,team-chameleon-2024,olmo-2-2024,qiu-why-2025},
with always-on query--key-side alternatives where it does not apply \citep{anson-controlling-2025}; loss
scaling and rollback alter the training procedure rather than a
location~\citep{micikevicius-mixed-2018,chowdhery-palm-2022}; magnitude controllers continuously clamp
gradients~\citep{zclip-2025,spam-2025,adagc-2025}, adjust learning rates using spectral
norms~\mbox{\citep{qi-taming-2025}}, or reparameterize weights to control spectral
norm~\citep{zhai-stabilizing-2023}; query--key reparameterization against weight oscillation in low-bit
quantization~\citep{liu-oscillationfree-2023} is a further QK-side alternative. Detectors report without acting
\citep{huang-mechanismdriven-2026,qiu-spectral-2025}, while prediction anticipates faults rather than
answering them~\citep{wortsman-smallscale-2023,emadi-rankaware-2026}. Native-hardware studies report the
instability and leave the channel-level account open~\citep{cim-pretraining-2026,dong-dissecting-2026}.
\paragraph{Older forms of the principle.} The broader principle that temporally correlated gradient estimates
destabilize training is older still~\citep{molybog-theory-2023}, though not in terms of attention or of
arithmetic error; prior work likewise leaves aggregate deviation's link to instability
open~\citep{golden-flash-2024}.
\paragraph{Why the streaming recurrence went unmodelled (\S2).} Low-precision accumulation was known to
destabilize training \citep{sakr-accumulation-2019}, but classical analysis expected shorter tiled reductions
to be safer \citep{higham-accuracy-1993}, leaving this failure mode unmodelled.

\section{Further Analyses}
\label{app:boundaries}

\paragraph{Why no per-step summary reads the gate.}
The per-step error summaries we registered in advance cannot tell sign structure apart. Take the
realized weight-gradient error of a collapsing run and scramble its element signs: the result agrees
with the original to three decimals on both, \NEdiagFoilLTwoRelDwq{} against
\NEdiagReimplLTwoRelDwq{} in relative \(\ell_2\) and \NEdiagFoilLinfDwq{} against
\NEdiagReimplLinfDwq{} in \(\ell_\infty\), while its per-head DC magnitude is the larger of the two.
The constructed foil shows that two errors of opposite sign structure can agree exactly on the summaries, and that the one
summary that does move between them moves the wrong way~\ifhmode\unskip\fi. Since
these are per-step readings of the error itself, and the property the outcomes turn on is a relation
between steps, this is less a gap in the particular summaries than a hint about where to look. That
prior work reads such aggregate proxies~\citep{golden-flash-2024} while leaving their link to
instability explicitly open is what motivated the matched design here.

\paragraph{The pathwise measurement.}
The discriminating quantity is pathwise. We fork the run live from one mid-trajectory state at which
the registered cross-step summaries all sit at their null values, and drive each fork with its own
variant. Write \(C_v\) for the norm of the accumulated per-head query--key contribution divided by
the sum of its per-step norms, at the layer adjacent to the runaway: a fork whose contributions
cancel over time sits near the value obtained by flipping the same series' signs at random, and a
fork that accumulates does not. The two arms that go on to collapse reach \NForkActualCv{} and
\NForkHeadonlyCv{} against sign-flipped nulls of \NForkActualCvNull{} and \NForkHeadonlyCvNull{},
and reproduce the \(\sigma(W^Q)\) signature at \NForkActualMaxSigma{} and \NForkHeadonlyMaxSigma{}.
The scrambled fork cancels, at \NForkScramCv{} against its own null, and survives to
\NForkScramStableTo{} steps; the mean-removed fork survives as well, at \NForkHeadcorrCv{}, near its own
null~\ifhmode\unskip\fi.

\paragraph{The cross-hardware portability check in full.}
Figure~\ref{fig:b200} repeats the dissociation on a single Blackwell B200 running the same
training program that produces the cluster collapses, with the same software-emulated quantizer.
The query--key-source run collapses at \NBtwohundredQkNofixCrash{} steps, with matched-step spectral
magnitudes agreeing with its cluster twin.

\ifdefined\arxivlayout\FloatBarrier\fi

\begin{figure*}[t]
\centering
\includegraphics[width=\textwidth]{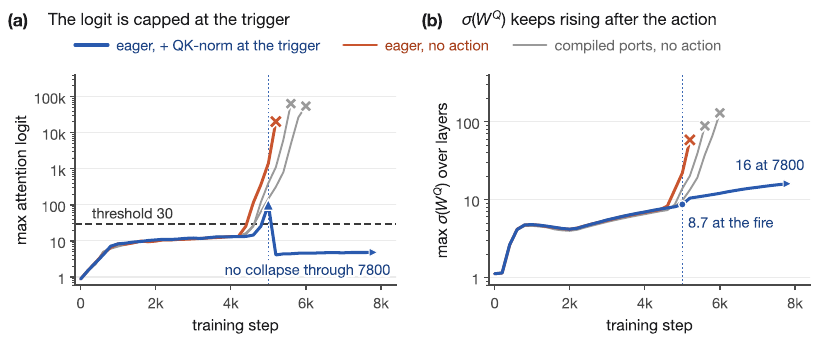}
\caption{The B200 portability check: one B200 running our exact code, one seed per arm.
\textbf{(a)}~Max attention logit. The compiled \(QK\)- and MLP-source arms (gray) collapse
(\(\times\)); in the matched eager pair the untreated twin collapses while the rescued run's
logit is capped mid-run at the trigger (\(\blacktriangle\)) --- the threshold carried over
from the H20 protocol unchanged. \textbf{(b)}~Under that rescue \(\sigma(W^Q)\) keeps rising:
the intervention bounds the logit, not the spectral norm. Traces stop at their collapse step,
so plotted height is not a severity; \(\blacktriangleright\) marks the scheduled horizon.}
\label{fig:b200}
\end{figure*}

The decisive arm moves the fault off attention on that same device. With the quantizer confined to
the two MLP weight gradients and the query--key gradients left unquantized, the run still
develops the query--key runaway and still collapses, at \NBtwohundredMlpNofixCrash{} steps. Over
the window before the first collapse the two sources' log spectral trajectories correlate at
\NBtwohundredMlpQkLogsigmaCorr{}. Triggering per-head query--key normalization mid-run then holds that same
faulted run healthy to \NBtwohundredMlpRescueStableTo{} steps at
\NBtwohundredMlpRescueFinalVal{} validation loss while \(\sigma(W^Q)\) keeps climbing to
\NBtwohundredMlpRescueMaxSigma{}. The MLP source is never repaired and \(\sigma(W^Q)\) never falls:
what the intervention acts on is the attention logit, and bounding that breaks the chain from
spectral growth to loss at its middle stage, leaving the growth to continue far more slowly than in
the untreated twin (Figure~\ref{fig:b200}). An otherwise identical run under the same
execution mode, without the intervention, collapses at \NBtwohundredMlpNofixEagerCrash{} steps --
after the intervention had already fired in its twin -- so the survival is attributable to the
action and not to how the run was executed.

\paragraph{The dose--response ordering behind \S2.}
The injected-error continuum referenced there orders as printed: crash steps
\NContinuumInjLamFourZeroZeroSZeroCrash{}, \NContinuumInjLamTwoZeroZeroSZeroCrash{},
\NContinuumInjLongBwdSZeroCrash{}, and \NContinuumInjLamZeroFiveZeroXTwentykSZeroCrash{} as the
magnitude falls from \(\lambda=4\) to \(\lambda=0.5\), one seed per rung, the last on an extended 20k schedule;
every tested rung ends in a crash.

\paragraph{Where fp32 accumulation is the boundary.}
Quantizing both operands of all three attention GEMMs to the same four-bit grid, with fp32
accumulation retained, leaves training healthy: neither of two seeds collapses through
\NEToENofixStableTo{} steps, with the registered seed's \(\sigma(W^Q)\) no higher than
\NEToENofixMaxSigma{} (a milder effective dose than the weight-gradient arms).
Four-bit arithmetic is therefore not by itself the trigger within this horizon: the observed
boundary lies between operand-level three-GEMM MXFP4 with fp32 accumulation and post-reduction
weight-gradient quantization.

\paragraph{The runaway across data-parallel shapes.}
At the eight-rank data-parallel shape, the unstripped weight-gradient references develop the full
\(\sigma(W^Q)\) runaway --- to \NDyadRefAllExtMaxSigma{} (model-wide) and \NDyadRefQkExtMaxSigma{}
(query--key) by their 20k horizons --- without loss-collapsing inside them, and on four GPUs the
MLP-source fault of \S3.2 likewise did not collapse within its tested horizon. The runaway, not the
crash step, is what transfers across shapes, so the main text's loss-crash statements are made at the
two-rank shape that exhibits them.
At a fixed global accumulation count the rank count also sets the distinct-token batch, inversely
(Appendix~\ref{app:setup}): the eight-rank arms sample a quarter as many distinct tokens per optimizer
step as the two-rank arms, so step counts on the two shapes are not directly comparable.

\paragraph{What the two fault sources share, and what they do not.}
Across the two weight-gradient sources, the runaway's dominant directions overlap only partially:
the squared-cosine overlap between the sources' leading runaway directions is
\NCapModeidSZeroOverlap{} and \NCapModeidSOneOverlap{} over the two seeds' established-runaway
windows --- far from identity, far above chance for random directions at this width. What the
sources share is a state-dependent query--key-local channel, not one fixed subspace; this is the
analysis behind the closing sentence of \S4.2.

\end{document}